\documentclass[nohyperref]{article}

\usepackage{amsmath,amsfonts,bm}

\def\eqref#1{equation~\ref{#1}}

\def\1{\bm{1}}

\def\mG{{\bm{G}}}

\DeclareMathAlphabet{\mathsfit}{\encodingdefault}{\sfdefault}{m}{sl}
\SetMathAlphabet{\mathsfit}{bold}{\encodingdefault}{\sfdefault}{bx}{n}
\newcommand{\tens}[1]{\bm{\mathsfit{#1}}}

\def\tM{{\tens{M}}}

\DeclareMathOperator*{\argmax}{arg\,max}
\DeclareMathOperator*{\argmin}{arg\,min}

\usepackage{microtype}
\usepackage{graphicx}
\usepackage{subfigure}
\usepackage{booktabs} 

\usepackage{hyperref}

\usepackage[accepted]{icml2023}

\usepackage{amsmath}
\usepackage{amssymb}
\usepackage{mathtools}
\usepackage{amsthm}
\usepackage{verbatim}
\usepackage{soul}

\usepackage[capitalize,noabbrev]{cleveref}

\theoremstyle{plain}
\newtheorem{theorem}{Theorem}[section]
\newtheorem{proposition}[theorem]{Proposition}

\theoremstyle{definition}
\newtheorem{definition}[theorem]{Definition}

\theoremstyle{remark}

\newtheorem*{lemma*}{Lemma}
\newtheorem*{theorem*}{Theorem}

\usepackage[textsize=tiny]{todonotes}

\usepackage{scrextend}
\icmltitlerunning{A Game-Theoretic Framework for Managing Risk in Multi-Agent Systems\hfill\thepage}

\begin{document}

\twocolumn[
\icmltitle{A Game-Theoretic Framework for Managing Risk in Multi-Agent Systems}



\icmlsetsymbol{equal}{*}

\begin{icmlauthorlist}
\icmlauthor{Oliver Slumbers}{ucl,hua}
\icmlauthor{David Henry Mguni}{hua}
\icmlauthor{Stephen Marcus McAleer$\,^4$}{}
\icmlauthor{Stefano B. Blumberg}{ucl}
\icmlauthor{Yaodong Yang}{pek}
\icmlauthor{Jun Wang}{ucl}
\end{icmlauthorlist}

\icmlaffiliation{ucl}{University College London, London, UK}
\icmlaffiliation{pek}{Peking University, Beijing, China}
\icmlaffiliation{hua}{Huawei Technologies, London, UK}

\icmlcorrespondingauthor{Oliver Slumbers}{oliver.slumbers.19@ucl.ac.uk}

\icmlkeywords{Machine Learning, ICML}

\vskip 0.3in
]



\printAffiliationsAndNotice{} 

\begin{abstract}
In order for agents in multi-agent systems (MAS) to be safe, they need to take into account the risks posed by the actions of other agents. However, the dominant paradigm in game theory (GT) assumes that agents are not affected by risk from other agents and only strive to maximise their expected utility.
For example, in hybrid human-AI driving systems, it is necessary to limit large deviations in reward resulting from car crashes. 
Although there are equilibrium concepts in game theory that take into account risk aversion, they either assume that agents are risk-neutral with respect to the uncertainty caused by the actions of other agents, or they are not guaranteed to exist.
We introduce a new GT-based Risk-Averse Equilibrium (RAE) that always produces a solution that minimises the potential variance in reward accounting for the strategy of other agents. Theoretically and empirically, we show RAE shares many properties with a Nash Equilibrium (NE), establishing convergence properties and generalising to risk-dominant NE in certain cases. To tackle large-scale problems, we extend RAE to the PSRO multi-agent reinforcement learning (MARL) framework. We empirically demonstrate the minimum reward variance benefits of RAE in matrix games with high-risk outcomes. Results on MARL experiments show RAE generalises to risk-dominant NE in a trust dilemma game and that it reduces instances of crashing by 7x in an autonomous driving setting versus the best performing baseline. 

\end{abstract}

\section{Introduction}
\label{introduction}

\begin{figure*}[ht!]
\centering
\includegraphics[width=0.85\linewidth]{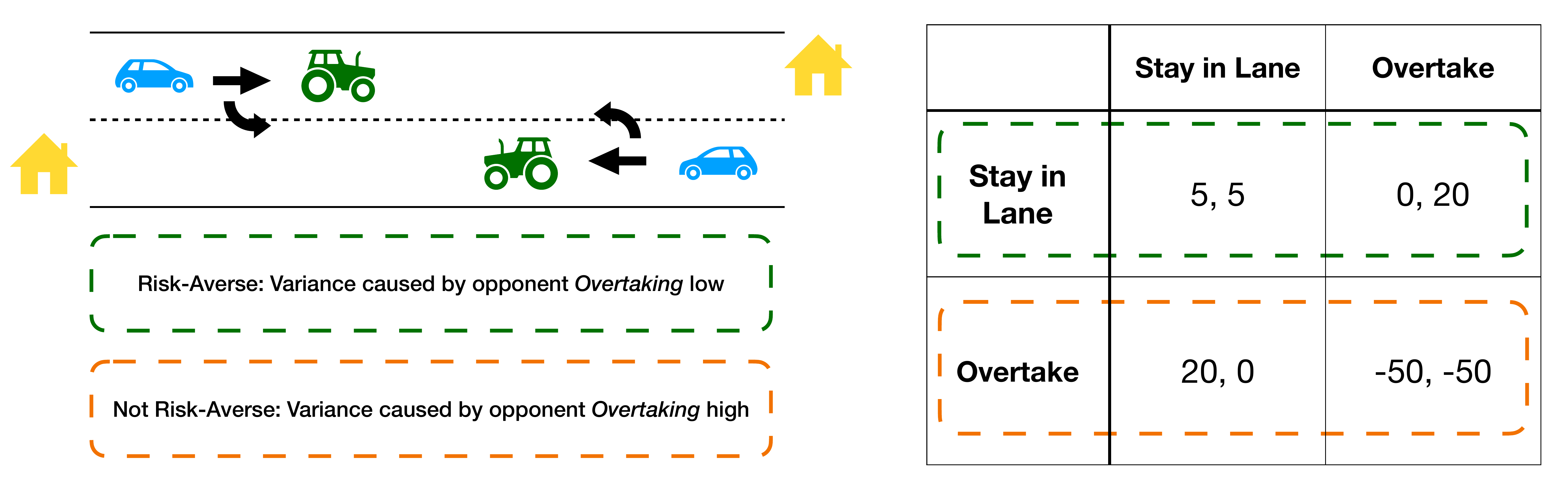}
\caption{Two cars are rewarded for reaching their destination quickly. They are stuck behind slow tractors but can stay in their lanes and arrive slowly. They can also overtake the tractors to arrive quickly, but if both overtake they will cause large delays, leading to large negative payoffs.}
\label{fig:self-drive}
\end{figure*}

Game Theory (GT) is a fundamental tool for  resolving problems within multi-agent systems (MAS) \cite{wellman2006methods}. Formalising scenarios in MAS as games allows practitioners to model stable outcomes in real-world settings by computing equilibrium solutions.
A core challenge for MAS research is building systems that perform effectively while mitigating the risk of adverse events for the agents within the system. In particular, as the level of integration with humans and AI agents increases, it becomes increasingly important to avoid dangerous events for humans i.e. car crashes \cite{gal2022multi}. In AI-only systems, risk can come in the form of agents taking low probability yet dangerous strategies (e.g. $\epsilon$-greedy policies outside of training \cite{mnih2015human}). Human-AI MAS differ markedly from those of simulated systems that involve computerised agents only, however the problem remains the same -- humans are prone to execution errors, a trait that seldom applies to computerised agents \cite{gal2022multi}. Tackling this challenge requires equilibrium concepts that can account for all possible rewards dependent on other agents' actions and, enable agents to adopt risk-averse strategies in response. 


There are two prominent approaches in GT that propose equilibrium concepts that address risk: Trembling Hand Perfect Equilibrium (THPE) \cite{bielefeld1988reexamination} and Quantal Response Equilibrium (QRE) \cite{mckelvey1995quantal}). However, due to the linearity of expected utility (EU) these concepts can undervalue comparatively large costs with low probability given an agent's tolerance for risk, which undermines their ability to successfully resolve notions of safety and risk in various practical MAS (e.g. crashing has a large negative utility). For example, as demonstrated in Fig.\ref{fig:self-drive}, a probability of 0.01\% of Player 2 \textit{Overtaking} only impacts Player 1's EU by 0.5. Therefore, based on EU, \textit{Overtaking} is preferable for Player 1, however exposing it to the possibility of congestion in the middle of the road. Furthermore, it is difficult to control the level of risk-aversion in these frameworks as they are defined only in terms of EU.


Other approaches characterise risk in terms of concave/convex utility functions and study convergence to classical GT equilibria. In \citealt{fiat2010players}, different attitudes to risk are theoretically considered, in particular a risk-averse variant based on including variance in the utility function similar to our proposal. However, they find that almost all of their risk-adjusted games will not have mixed Nash equilibrium. Even if the risk adjusted game does have a mixed Nash equilibrium, small mistakes made by players in interpreting the payoffs of the game will likely cause the equilibrium to be unstable. In contrast, our approach guarantees the existence of a risk-adjusted equilibrium.
Furthermore, \citealt{fiat2010players} only provide theoretical analyses and we are interested in practical solvers to attain a solution.

In this paper we propose a novel equilibrium concept called \textit{Risk-Averse Equilibrium} (RAE). RAE distinguishes itself from THPE and QRE by including the second moment (variance) of the utility function with respect to the strategies of other agents. Unlike these approaches RAE has a risk-aversion parameter explicitly controlling the amount of risk an agent is willing to accept. This places larger emphasis on deviations from expected utility, where variance in RAE values the 0.01\% \textit{Overtake} probability in Fig. \ref{fig:self-drive} at 47.6$\gamma$ (vs. 0.5), where $\gamma > 0$ and generally at minimum 0.1, see Appendix \ref{appendix:hyperparams}. We show that RAE can be computed using numerical and iterative methods, unlike the theoretical approaches in \cite{fiat2010players}, which unlock its ability to resolve modelling large-scale MARL games simulating real-world settings. We also show that, for any desired expected utility, RAE minimises the corresponding variance. In other words,
RAE reduces highly adverse outcomes caused by the actions of other agents in the system.


To demonstrate the benefits of RAE, we perform a series of experiments in risky matrix games, and two larger-scale MARL-problems which are closer to real-world scenarios: a grid-world Stag-Hunt trust dilemma, and an autonomous driving scenario. Our results validate our theoretical advances on risky matrix games by showing that the RAE solutions provide the same EU as our baselines, whilst being safer in providing lower variance. In the Stag-Hunt MARL setting, RAE outperforms the baselines and arrives at the safe solution but the other methods do not. In the autonomous driving setting, RAE reduces crashes 7x in test episodes as compared with the best performing baseline equilibrium.

\textbf{Our contributions are:}
\begin{enumerate}
    \item We propose RAE, a new solution concept 
    that accounts for expected utility (EU) and utility variance (UVar) in determining a strategy (Sec. \ref{sec:rae}).
    \item We prove that RAE's strategy is the minimum variance solution (Prop. \ref{prop:min-var-solution}) and we prove RAE's existence (Thrm. \ref{thrm:existence}).
    \item We introduce two solvers to compute RAE in small action spaces (Sec \ref{sec:solvers-sfp}) and large action spaces (Sec. \ref{sec:solvers-psro}). 
    \item We validate RAE on risky matrix games and in experiments on two MARL settings: a grid-world Stag-Hunt (Sec. \ref{sec:exps-stag}) and autonomous driving scenario (Sec. \ref{sec:exps-drive}), where RAE outperforms state-of-the-art GT approaches THPE, QRE and simple NE.
\end{enumerate}

\section{Related Work}
\label{sec:related_work}

\textbf{Risk in GT: Baselines} 
  \citealt{harsanyi1988general} introduced risk-dominant Nash equilibria (NE) \citep{nash1951non}, which, when the strategies of other agents is unknown, leads to the NE with the lowest losses if deviated from. However, if none of the NE are robust to risk initially, then the risk-dominant strategy amongst NEs also will not be. THPE \cite{bielefeld1988reexamination} models risk by having agents 'tremble' through all actions having positive probability in a mixed-strategy. However, iterated deletion of weakly dominated strategies can lead to the THPE strategy being removed from consideration. \citealt{mckelvey1995quantal} introduce QRE which accounts for potential errors in strategy selection to more accurately represent human observed strategies. Note, THPE and QRE utilise EU as their risk measure which is insensitive to low probability, high-value deviations as long as they are offset by high probability mean values \cite{royset2022risk}. THPE, QRE and NE are baselines. 

  \textbf{Risk in GT: Other Approaches}
  \citealt{yekkehkhany2020riskaverse} propose a mean-variance equilibrium where the variance relates to reward probabilities, rather than variance caused by the strategies of others as in RAE. Notably, in the model-free machine learning setting reward probabilities are generally not known, and is not within the scope of this work. \citealt{fiat2010players} consider how NE existence is impacted by non-EU maximising agents, and derive results for multiple risk categorisations, however limit their work to theoretical propositions that do not extend to baselines for this work. Risk in competitive network games \cite{wardrop1952road} is widely studied, and is based on a generalisation of the classical selfish routing 'game' \cite{beckmann1956studies} to incorporate uncertain delays. The impact of mean-risk utility frameworks on WE is particularly well studied \cite{ordonez2010wardrop, lianeas2019risk, cherukuri2019sample}, however WE is not applicable to non-network games as studied in this work. In terms of NE, the major distinction in risk analysis is between non-atomic (i.e. infinite agents) and atomic (i.e. finite agents) settings. In non-atomic settings \citealt{meir2015congestion, nikolova2012stochastic} the marginal impact in terms of risk of each agent on each other is infinitesimal and this leads to a distinctly different analysis than required in the atomic setting of this work \cite{nikolova2014mean}. In the atomic setting which parallels more with our setting, \citealt{nikolova2014mean} study a mean-standard deviation model of travel time along a network path, and whilst they are able to show the existence of pure-strategy NE in exogenous risk settings, they are unable to in endogenous risk settings concerned in our work. \citealt{piliouras2016risk} also study a setting where the risk is largely exogenous, as it is defined by randomised schedulers which controls the ordering through congestion edges in the network, which is not a concept in our setting and therefore can not act as a baseline.

\textbf{Risk in MARL:} RAE fits broadly into risk-sensitive MARL solutions such as: RMIX \citep{qiu2021rmix} which optimises decentralised CVaR policies in cooperative risk-sensitive settings, RAM-Q and RA3-Q \citep{gao2021robust} utilise an adversarial approach to promote variance reduction, or risk-sensitive DAPG \citep{eriksson2022risk} which approaches risk in Bayesian games in terms of the CVaR induced by the possible combinations of types in the game. However, as we are specifically concerned with GT-based equilibrium concepts we will not directly compare to these methods.


\textbf{GT and MARL} GT and RL have overlapped in settings where the number of actions in a game becomes too big to write down trivially. For these games with larger action spaces, we consider Policy-Space Response Oracles (PSRO) \cite{lanctot2017unified, mcaleer2020pipeline, pmlr-v139-perez-nieves21a, feng2021neural, mcaleer2022self} which generalises the Double Oracle (DO) \cite{mcmahan2003planning, dinh2021online, mcaleer2021xdo} framework from small action spaces to large action spaces by replacing actions with RL policies. In this work, we propose a framework at the overlap between risk-averse GT and risk-averse MARL which involves adaptations to the risk-neutral frameworks mentioned here. In future work we will investigate applying other recent deep RL approaches for finding equilibria~\citep{perolat2022mastering, mcaleer2022escher, sokota2022unified} to our solution concept.

\section{Risk-Averse Equilibrium Framework}
\label{sec:rae}
We introduce RAE and detail its derivation from a mean-variance utility function. We derive key properties of RAE which characterise the two following important benefits: \textbf{1.} RAE is a minimum risk solution \textbf{2.} The existence of RAE for any finite N-player game (under standard conditions).


\subsection{RAE Derivation}
\label{sec:rae-derivation}


We begin by describing the underlying formalism of a normal form game (NFG). An NFG is the standard representation of strategic interaction in GT. A finite $n$-person NFG is a tuple $(N,A, u)$, where $N$ is a finite set of $n$ players, $A = A^1 \times, ..., \times A^n$ is the joint action profile, with $A^i$ being the actions available to player $i$, and $u = (u^1, ..., u^n)$ where $u^i: A \rightarrow \mathbb{R}$ is the real-valued utility function for each player. A player plays a mixed-strategy, $\boldsymbol{\sigma}^i \in \Delta_{A^i}$, which is a probability distribution over their possible actions. In Sec. \ref{sec:solvers-psro} we replace atomic actions with neural networks and will therefore re-define our notation to keep clarity between the two game schemes.

The objective of RAE is to provide an equilibrium solution that is robust to other agents taking \textit{any} action. Our approach considers both the EU (mean) and the potential UVar caused by an opponent's strategy, in particular noting that \textit{all} actions may be taken by the opponent as if mistakes (low probability actions) can happen. Whilst low probability actions are not literally mistakes, they are simply a technical device to mimic the idea of a mistake, as described in \citealt{bielefeld1988reexamination}.

For simplicity, we provide definitions based on playing a symmetric game, such that two players share an action set $\mathcal{A}$, and a utility function $u$. We extend this to the non-symmetric case in Appendix \ref{appendix:asymmetric-utility}. Define the utility of action $a_i \in \mathcal{A}$ against action $a_j \in \mathcal{A}$ as $u(a_i, a_j)$ and the full utility matrix as $\tM$, where the entry $\tM_{i,j}$ refers to $u(a_i, a_j)$ and $\tM_{i}$ refers to $u(a_i, a_j) \; \forall j $, i.e. the vector of utilities that action $a_i$ receives against all other actions. We now define the \textit{expected} utility of the mixed-strategy for player 1 $\boldsymbol{\sigma}$ versus the mixed strategy for player 2 $\boldsymbol{\varsigma}$ as

\begin{equation}
\begin{aligned}
    u(\boldsymbol{\sigma}, \boldsymbol{\varsigma}, \tM) &= \sum_{a_i \in A} \sum_{a_j \in A} \sigma(a_i) \varsigma(a_j) u(a_i, a_j) \\ &= \boldsymbol{\sigma}^{T}\cdot \tM \cdot \boldsymbol{\varsigma}.
\end{aligned}
\end{equation}

The weighted co-variance matrix for $\tM$ (i.e. the UVar values) is a $|A| \times |A|$ matrix $\boldsymbol{\Sigma}_{\tM, \boldsymbol{\varsigma}}$ = $[c_{ij}]$ with entries 
\begin{equation}
\begin{aligned}
\label{eq:covariance}
c_{jk} = \sum_{a_i \in A} \varsigma(a_i) \big(u(a_i,a_j)-\bar{\tM}_j\big)\big(u(a_i, a_k) - \bar{\tM}_k\big),
\end{aligned}
\end{equation}
where $\bar{\tM}_i = \sum_{k=1}^{|A|} \varsigma_k u(a_i, a_k)$ is the EU for action $i$ given the opponent mixed-strategy $\boldsymbol{\varsigma}$. This is a standard co-variance matrix, but the entries are weighted by the likelihood of them being selected by the opponent. A uniform weighting could be used, however we hypothesise that in terms of minimising UVar it is more appropriate to hedge against the variance caused by higher likelihood actions. Due to the nature of Eq. \ref{eq:br-map}, all actions will, by design, receive positive probability $(> \epsilon)$ under our framework and therefore will always provide some weight in the variance calculation, leading to low likelihood high-variance actions still having a large impact. This accounts for the central concept of RAE, that safe play should account for all actions. This allows us to define the mixed-strategy $\boldsymbol{\sigma}$ UVar as:
\begin{equation}
\begin{aligned}\label{eq:var}
\operatorname{Var}(\boldsymbol{\sigma}, \boldsymbol{\varsigma}, \tM) &= \sum_{k=1}^{|A|} \sum_{n=1}^{|A|} \sigma(a_k) \sigma(a_n) c_{kn} \\ 
&= \boldsymbol{\sigma}^{T} \cdot \boldsymbol{\Sigma}_{\tM, \boldsymbol{\varsigma}} \cdot \boldsymbol{\sigma}.
\end{aligned}
\end{equation}
The total utility function $r$ which considers EU and UVar for mixed-strategy $\boldsymbol{\sigma}$ is,
\begin{align}\label{eq:util-func}
    r(\boldsymbol{\sigma}, \boldsymbol{\varsigma}, \tM) = u(\boldsymbol{\sigma}, \boldsymbol{\varsigma}, \tM) - \gamma \operatorname{Var}(\boldsymbol{\sigma}, \boldsymbol{\varsigma}, \tM),
\end{align}
where $\gamma \in \mathbb{R}$ is the risk-aversion parameter. 

Applying Eq. (\ref{eq:util-func}) to Fig. (\ref{fig:self-drive}) we show why the utility function \ref{eq:util-func} is desirable. Consider two joint strategy profiles, $\boldsymbol{S}_1 = ((1-\epsilon, 0+\epsilon), (1-\epsilon, 0+\epsilon))$ and a THPE $\boldsymbol{S}_2 = ((0+\epsilon, 1-\epsilon), (1-\epsilon, 0+\epsilon))$ where $(1 - \epsilon, 0+\epsilon)$ represents playing \textit{Stay in Lane} with probability $(1 - \epsilon)$. $\epsilon = 0.01$ for this example. Profile $\boldsymbol{S}_1$ receives $u(\boldsymbol{S}_1) \approx 5$ and the THPE profile receives $u(\boldsymbol{S}_2) \approx 20$. However, $\operatorname{Var}(\boldsymbol{S}_1) = 0.32$ and $\operatorname{Var}(\boldsymbol{S}_2) = 47.6$, i.e. the THPE strategy has huge variance for Player 1. Therefore, $r(\boldsymbol{S}_1) \approx 5 - 0.32\gamma$ and $r(\boldsymbol{S}_2) \approx 20 - 47.6\gamma$ and we have for any risk-aversion parameter $\gamma > 0.32$ it is optimal to play $\boldsymbol{S}_1$.

To define RAE, we first define the best-response map:
\begin{equation} \label{eq:br-map}
\begin{aligned}
\boldsymbol{\sigma}^*(\boldsymbol{\varsigma}) \in \argmax_{\boldsymbol{\sigma}}r(\boldsymbol{\sigma}, \boldsymbol{\varsigma}, \tM) \\
\text{s.t. } \sigma(a) \geq \epsilon  \text{  }, \forall a \in A\\
\boldsymbol{\sigma}^{T}\textbf{1} = 1, 
\end{aligned}
\end{equation}
where due to the quadratic term $\boldsymbol{\sigma}^{T} \cdot \Sigma_{\tM, \boldsymbol{\varsigma}} \cdot \boldsymbol{\sigma}$ and the constraints $(\epsilon > 0)$, we have a Quadratic Programming (QP) problem. The programme finds $\boldsymbol{\sigma}^*$ such that the total utility is maximised, whilst ensuring no actions are assigned negative probability, and that the probabilities sum to one. Finally, based on Eq.  (\ref{eq:br-map}) we are able to define RAE:

\begin{definition}[RAE]\label{rae_def} A strategy profile $\{\boldsymbol{\sigma}, \boldsymbol{\varsigma}\} $ is a risk-averse equilibrium if both $\boldsymbol{\sigma}$ and $\boldsymbol{\varsigma}$ are risk-averse best responses, in that they satisfy Eq. \ref{eq:br-map}, to each other.
\end{definition}

\subsection{RAE Properties}
\label{sec:rae-properties}
The first property of RAE is that, given an opponent's strategy $\boldsymbol{\varsigma}$, one's owns strategy $\boldsymbol{\sigma}$ is the minimum UVar solution available given their desired EU $\mu_b$:

\begin{proposition}
\label{prop:min-var-solution}
Given $\boldsymbol{\varsigma}$ and any desired EU $\mu_b$, there exists a $\gamma$ such that the solution to Eq. \ref{eq:br-map} receives $\mu_b$ with the minimum possible UVar. 
\end{proposition}

Proof deferred to Appendix \ref{appendix:proofs-min-var-solution}. The consequence of this proposition matches with the goal of our paper. $\gamma$ is used to select how much EU $\mu_b$ is desired and, given the opponent's strategy $\boldsymbol{\varsigma}$, there is no other solution that can provide better risk performance than RAE. The main downside of this is that $\gamma$ is a hyper-parameter and it may be difficult to know prior to training how it will exactly match to $\mu_b$.


Secondly, a common property of most GT equilibrium is that a solution exists, at least in the finite game setting. For RAE, we note the following result in mixed-strategies:
\begin{theorem}\label{thrm:existence}
For any finite N-player game where each player $i$ has a finite $k$ number of pure strategies, $A^i = \{a^i_1, ..., a^i_k\}$, an RAE exists.
\end{theorem}
We defer the proof of the result to Appendix \ref{appendix:proofs-existence}. By establishing the existence of RAE solutions we have validated the practical relevance of RAE. 
\begin{figure*}[ht!]
\centering
\includegraphics[width=0.8\linewidth]{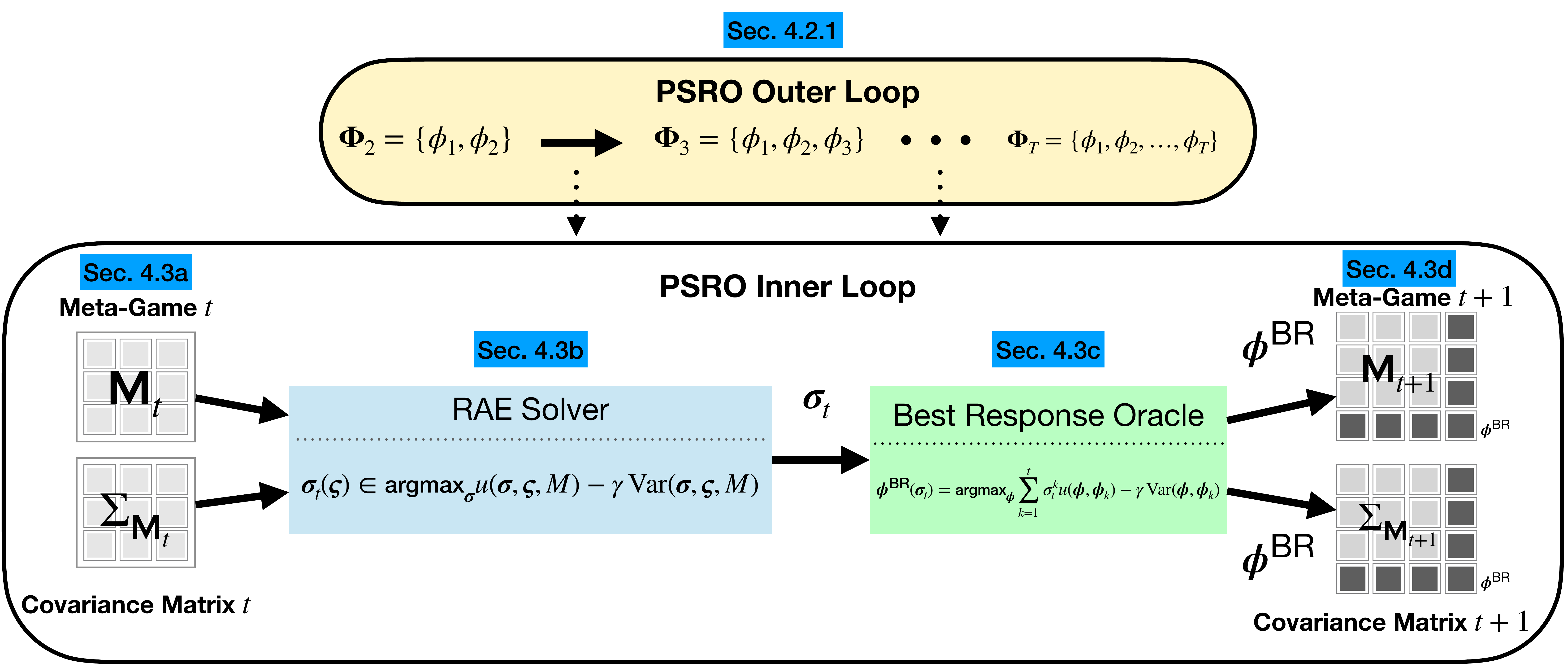}
\caption{PSRO-RAE. Each element points to a corresponding subsection in the text, denoted in blue boxes. Note that $u(\cdot)$ and $\operatorname{Var}(\cdot)$ are overloaded to represent utility/variance between distributions over a population or utility/variance between two policies.}
\label{fig:psro_vis}
\end{figure*}

\section{Finding RAE}
This section proposes two solvers to find RAE, with stochastic fictitious play (SFP) for small action-spaces in Sec. (\ref{sec:solvers-sfp}) and with PSRO-RAE for large action-spaces in Sec. \ref{sec:solvers-psro}.

\subsection{Stochastic Fictitious Play}
\label{sec:solvers-sfp}
We start by showing that our total utility function is a form of SFP \citep{fudenberg1993learning} which can find an RAE in small NFGs. SFP has convergence guarantees in a selection of games, e.g. potential games \citep{monderer1996fictitious, monderer1996potential} and finite two-player zero-sum games \citep{10.2307/1969530}. Furthermore, SFP is also robust empirically in terms of convergence in game classes \citep{goldberg2013approximation, ganzfried2020fictitious} not listed, and we mirror these observations in Appendix \ref{appendix:robust}. 

SFP is a learning process where players choose a best response to others time-average strategies. In SFP, a group of $n \geq 2$ players repeatedly play a $n-$player NFG. The state variable is $Z_t \in \Delta_S$, whose components $Z^i_t$ describe the time averages of each player's behaviour,
\begin{align*}
    Z^i_t = \frac{1}{t} \sum_{u=1}^t \boldsymbol{\sigma}^i_t
\end{align*}
where $\boldsymbol{\sigma}^i_t \in \Delta_{A^i}$ represents the observed strategy of player $i$ at time-step $t$. Each player best responds to the time-average strategy of their opponent, $Z_t^{-i}$, by maximising a perturbed utility function $\bar{u}$
\begin{align}\label{eq:sfp-br}
    \boldsymbol{\sigma}_{t+1}^i &= \argmax_{\boldsymbol{\sigma}}\bar{u} \\  &= \argmax_{\boldsymbol{\sigma}} u^i(\boldsymbol{\sigma}, Z_t^{-i}, \tM) - \lambda v^i(\boldsymbol{\sigma})  
\end{align}

where $v^i(\boldsymbol{\sigma}) : \Delta_{A} \rightarrow \mathbb{R}$ perturbs $u^i$ such that it is strictly concave (unique global maximum) whilst applying greater than zero probability to all actions.

\begin{proposition}
\label{prop:sfp-convergence}
Replacing the best-response Eq. (\ref{eq:sfp-br}) with the best-response map Eq. (\ref{eq:br-map}) satisfies the conditions of $\bar{u}^i$ for a SFP process.
\end{proposition}

Note, SFP does not necessarily converge in all game classes (but is robust empirically, see Appendix \ref{appendix:robust}). Therefore, we show that if the SFP process does converge to a strategy then that strategy is an RAE.

\begin{proposition}\label{prop:sfp}
Suppose the SFP sequence $\{Z_t\}$ converges to $\boldsymbol{\sigma}$ in observed strategies \footnote{Convergence in time-average $Z_t$ does not imply convergence in the actual strategy taken at each $t$, but may imply cyclic actual behaviour that results in average behaviour converging.}, then $\boldsymbol{\sigma}$ is a risk-averse equilibrium.
\end{proposition}
Note for SFP we require a stronger notion of convergence in observed strategies $\boldsymbol{\sigma}_t^i$ rather than in beliefs $Z_t^i$, but a converged final $\boldsymbol{\sigma}_t^i$ is a risk-averse equilibrium.

\subsection{PSRO-RAE}
\label{sec:solvers-psro}

   \begin{figure*}[ht!]
\centering
\includegraphics[width=0.9\linewidth]{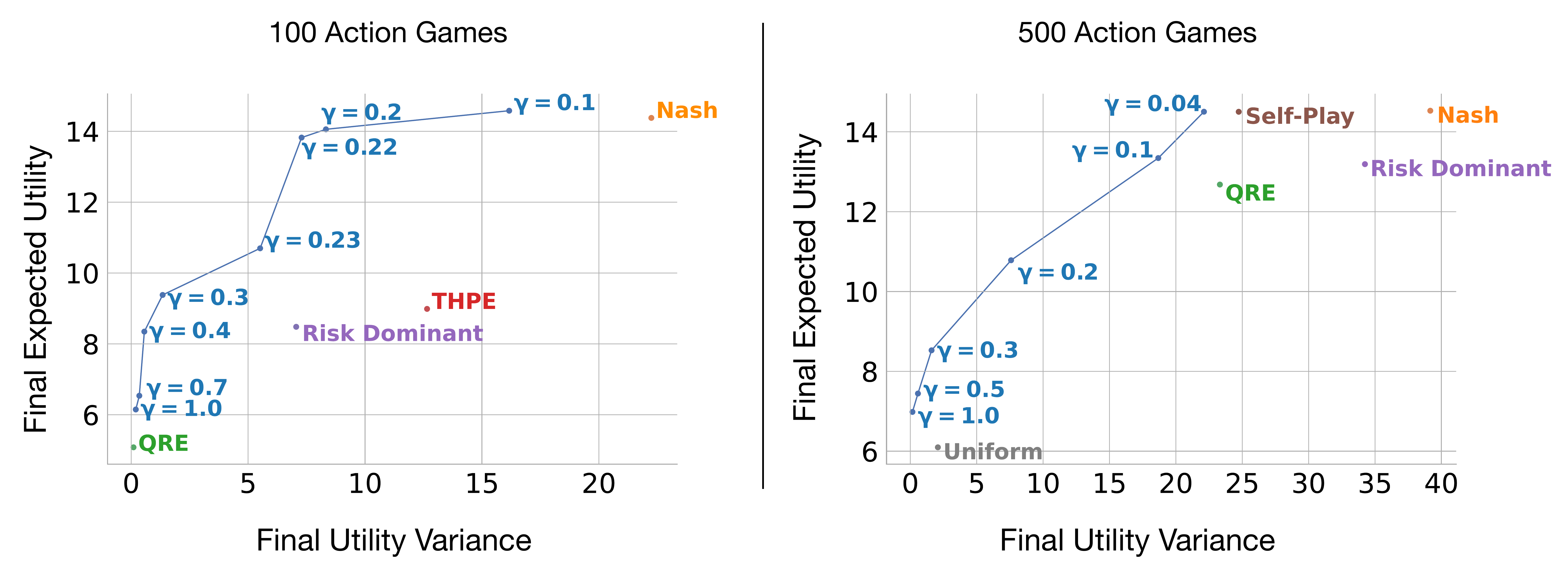}
\caption{a) SFP on NFGs with 100 actions, b) PSRO on NFGs with 500 actions. Final EU vs. UVar results. RAE values for multiple $\gamma$ form an \textit{efficient frontier} \cite{markowitz1991foundations} and show that, whilst baselines achieve similar EU they have large UVar solutions. In Fig. a) we exclude the payoff dominant NE result as its UVar was too large, whilst in Fig. b) we exclude the THPE result for the same reason.}
\label{fig:nfg}
\end{figure*}

For games that can't be displayed in the normal-form, we extend the iterative solution framework PSRO \cite{lanctot2017unified} to RAE, which uses RL policies as proxies for actions. PSRO-RAE approximates equilibria in large games by finding a small representative collection of risk-averse RL policies which are sampled by RAE. Whilst PSRO does not have any practical convergence guarantees (in the limit \textit{all} policies may be found and displayed in the normal-form, allowing for an exact solution), PSRO generally finds strong solutions without requiring all potential policies \cite{pmlr-v139-perez-nieves21a, feng2021neural, mcaleer2022anytime}. We provide a visualisation of the PSRO-RAE process in Fig. (\ref{fig:psro_vis}), and provide an algorithm in Appendix \ref{appendix:pseudo-code}. 

Consider two-player stochastic games $\mG$ defined by the tuple $\{\mathcal{S}, \mathcal{A}, \mathcal{P}, \mathcal{G}\}$, where $\mathcal{S}$ is the set of states, $\mathcal{A} = \mathcal{A}^1 \times \mathcal{A}^2$ is the joint action space, $\mathcal{P} : \mathcal{S} \times \mathcal{A} \times \mathcal{S} \rightarrow [0,1]$ is the state-transition function and $G = \{g_1, g_2\}$ is the set of rewards where $g^i:S \times \mathcal{A} \rightarrow \mathbb{R}$ is the reward function for player $i$ (we use reward for MARL settings, and utility for NFG settings). An \textit{agent} is a policy $\boldsymbol{\phi}$, where a policy is a mapping $\boldsymbol{\phi} : \mathcal{S} \times \mathcal{A} \rightarrow [0,1]$ which can be described in both a tabular form or as a neural network. The expected reward between two \textit{agents} is defined to be $M(\boldsymbol{\phi}_i, \boldsymbol{\phi}_j)$ (i.e., in the same manner defined for NFGs in Sec. \ref{sec:rae-derivation}), and represents the expected reward to agent $\phi_i$ against opponent $\phi_j$. 

\subsubsection{PSRO Outer Loop}

PSRO does $T \in \mathbb{N}^{+}$ updates on a meta-game $\tM$ (an NFG made up of RL policies as actions). At every iteration $t \leq T$, a \textit{player} is defined by a population of fixed \textit{policies} $\boldsymbol{\Phi}_{t} = \boldsymbol{\Phi}_{0} \cup \left\{\boldsymbol{\phi}_1, \boldsymbol{\phi}_2, ... , \boldsymbol{\phi}_{t}\right \}$, where $\boldsymbol{\Phi}_{0}$ is the initial random policy. For notation convenience, we consider the \emph{single-population} case where players share the same $\boldsymbol{\Phi}_t$, and refer the reader to Appendix \ref{appendix:asymmetric-psro} for the multi-population formulation. 

\subsubsection{PSRO Inner Loop}

\textbf{a, d) Meta-Game \& Covariance Matrix}
At the start of the iteration $t$ inner loop, each population has a \textit{meta-game} $\tM_t$, a reward matrix between all the \textit{policies} in the population, with individual entries $M(\boldsymbol{\phi}_i, \boldsymbol{\phi}_j) \; \forall \boldsymbol{\phi}_i, \boldsymbol{\phi}_j \in \boldsymbol{\Phi}_{t}.$ In addition, each population also generates a covariance matrix $\mathbf{\Sigma}_{\tM_t}$ defined by Eq. \ref{eq:covariance}. At the end of iteration $t$ inner loop, both $\tM_t$ and $\mathbf{\Sigma}_{\tM_t}$ are updated to include a new policy. 

\textbf{b) Meta Distribution}
We require a way to select which $\phi_t \in \boldsymbol{\Phi}_t$ will be used as training opponents. The function $f$ is a mapping $f : \tM_t \rightarrow [0,1]^t$ which takes as input a meta-game $\tM_t$ and outputs a \textit{meta-distribution} $\boldsymbol{\sigma}_t = f(\tM_t)$. The output $\boldsymbol{\sigma}_t$ is a probability assignment to each \textit{policy} in the population $\boldsymbol{\Phi}_t$, the equivalent of a mixed-strategy in a NFG, except actions are now RL policies. We apply RAE (Def. \ref{rae_def}) as the meta-solver. As $\boldsymbol{\phi}$ are RL policies then the policies are sampled by their respective probability in $\boldsymbol{\sigma}_t$.

\begin{figure*}[t!]
\centering
\includegraphics[width=0.8\linewidth]{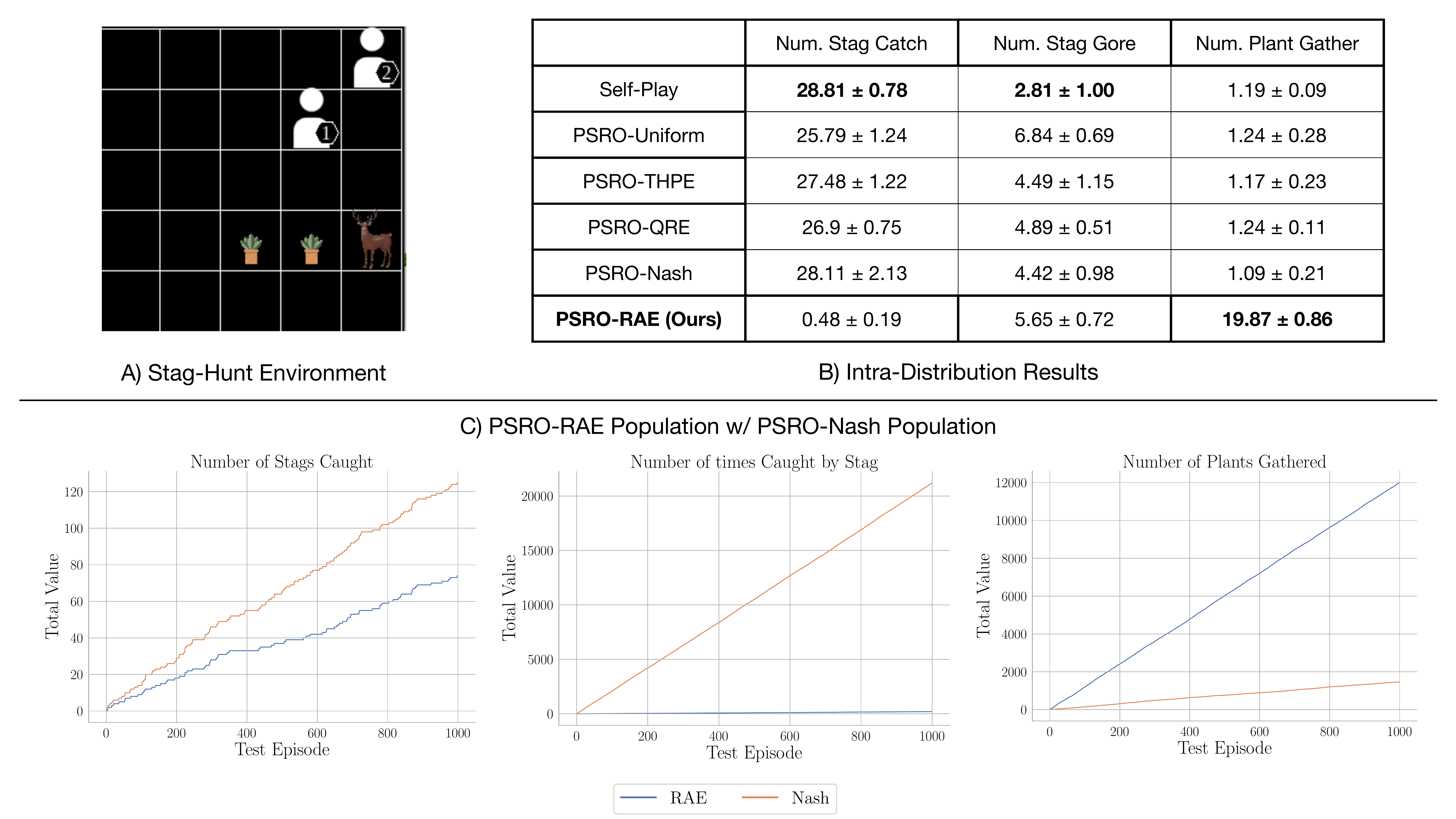}
\caption{Stag-hunt environment results. A) The environment B) Intra-distribution results, e.g. both agents are controlled by a PSRO-RAE population C) Cumulative results over 1000 test episodes for a PSRO-RAE population vs. PSRO-Nash population.}
\label{fig:stag_hunt}
\end{figure*}

\textbf{c) Best Response Oracle}
At each epoch $\boldsymbol{\Phi}_t$ is augmented with a new policy that is a \textit{best-response} (BR) to the meta-distribution $\boldsymbol{\sigma}_t$. The BR oracle aims to optimise the same objective function of that optimised by the meta-distribution. For example, in Vanilla PSRO, the Nash meta-distribution optimises for environment reward and the BR oracle also optimises a new agent in terms of EU. This can be found with any optimisation process such as RL or an evolutionary algorithm. In our setting the meta-distribution optimises two metrics, the EU and the UVar, and therefore, we need a BR oracle that optimises the same dual objective. 

In terms of RL quantities, this translates to maximising the expected total reward (i.e. the total of the per-step rewards) whilst minimising the variance of the total reward caused by the sampling of different RL agents from $\boldsymbol{\sigma}_t$. 

To achieve this, we follow the approach of \citep{zhang2020mean} that optimises both the total reward and per-step reward variance by solving an augmented MDP where the per-step reward $g^i_t$ is replaced by:

\begin{align*}
    \hat{g}^i_t = g^i_t - \lambda (g_t^i)^2 + (2 \lambda g_t^i y_i)
\end{align*}

where $y_i = \frac{1}{T}\sum_{t=1}^T g_t^i$ is the average of the per-step rewards during the data collection phase. 

We choose the per-step reward as it is an upper bound of the total reward variance \citep{bisi2019risk}, therefore reducing per-step variance will also reduce total variance, and is more effective computationally \citep{zhang2020mean}. Additionally, as this variance is also with respect to the sampling probability defined by $\boldsymbol{\sigma}_t$ this optimises the correct co-variance matrix which is also weighted by $\boldsymbol{\sigma}_t$. 

 \section{Experiments and Results}
 \label{sec:exps}

 We validate RAE through three questions: 1) Does RAE find lower UVar solutions than baselines? 2) Do RAE strategies overlap with risk-dominant NE in some scenarios? 3) Are RAE strategies safe in a safety-critical driving scenario? Full experimental details in Appendix \ref{appendix:envs}.

\subsection{Does RAE find lower UVar solutions than baselines?} 
 \label{sec:exps-nfgs}
\textbf{Motivation/Overview} Prop. \ref{prop:min-var-solution} shows RAE can find a spectrum of solutions encompassing many EU values, whilst minimising the corresponding UVar. Therefore, if RAE can match the baselines EU, whilst achieving lower UVar, then RAE is a better solution if safety is of concern.


\textbf{Baselines} NE (including risk/payoff dominant), THPE and QRE introduced in Sec. \ref{sec:related_work}.

\textbf{Experiment: Matrix Coordination Games} NFGs where some actions provide a high EU if other agents select them, but have large costs if not. Other actions have lower coordinated EU but lower costs. These games are designed to highlight the issues of focusing on EU and ignoring UVar.  


 \textbf{Results} Results in Fig. (\ref{fig:nfg}), where (A) represents games with 100 actions solved using SFP (Sec. \ref{sec:solvers-sfp}), and (B) represents games with 500 actions solved using PSRO (Sec. \ref{sec:solvers-psro}. We plot RAE solutions for multiple values of $\gamma$ to generate a theoretical \textit{efficient frontier}. An efficient frontier shows for values of EU what is the minimum possible UVar. Our results show that, whilst the baselines achieve a diverse range of EU values, they are unable to find the minimum UVar solution which RAE finds. This shows the strong flexibility of our approach, in that it is able to attain any EU that the baselines can achieve, whilst finding a lower UVar solution. This suggests that, if safety is of concern, then RAE is a better choice as any desired EU can be achieved whilst achieving a lower UVar than any of our baselines.

 \subsection{Are RAE strategies risk-dominant?} 
 \label{sec:exps-stag}
  \begin{figure*}[t!]
\centering
\includegraphics[width=0.85\linewidth]{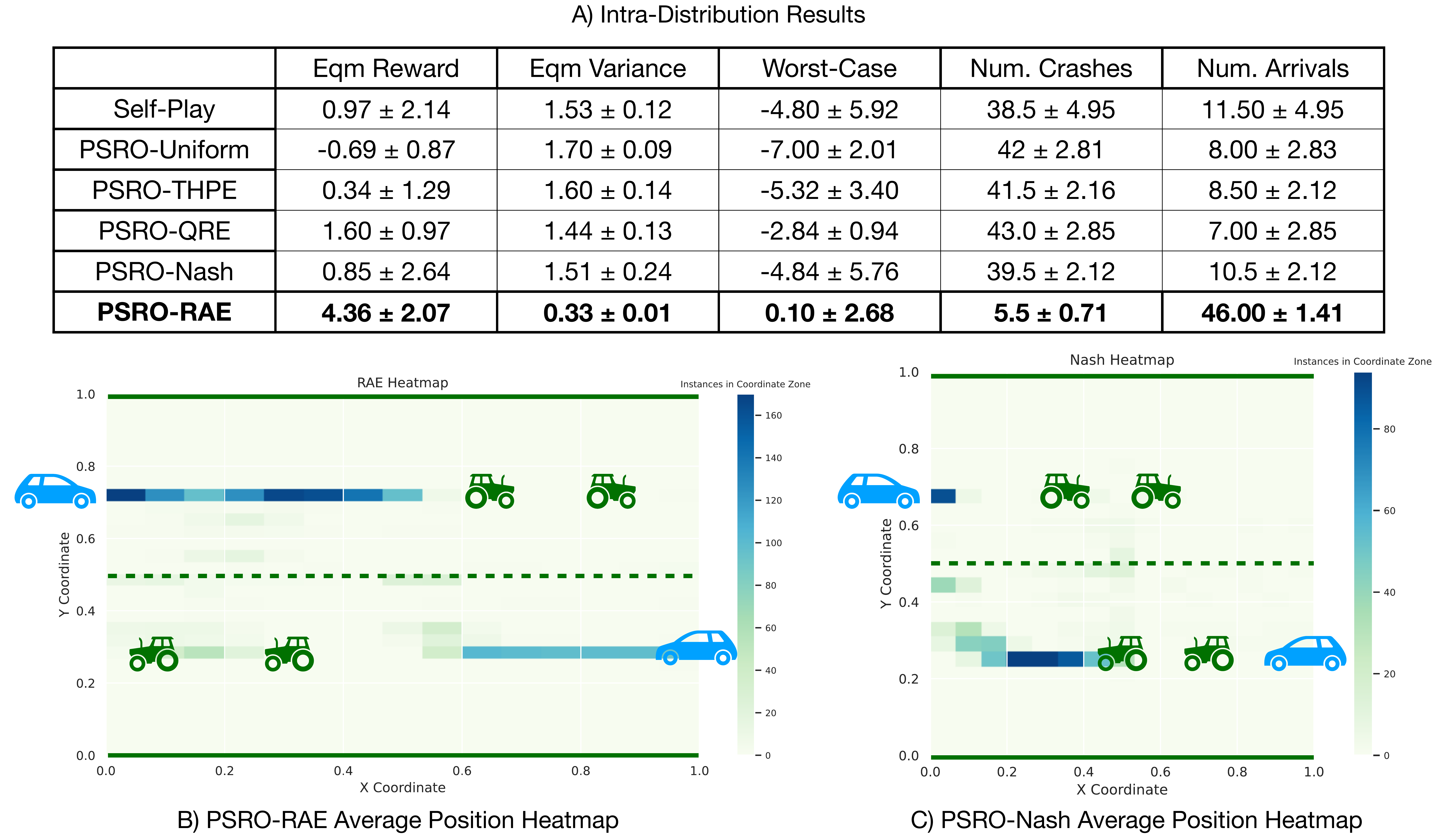}
\caption{A) Results on 50 episodes over 5 seeds for intra-distribution testing, e.g. both agents controlled by PSRO-RAE B) Average position heatmap for PSRO-RAE solution over 200 episodes C) Average position heatmap for PSRO-Nash solution over 200 episodes.}
\label{fig:drive_res}
\end{figure*}
 
 \textbf{Motivation} In some scenarios, it is likely that the only viable solutions overlap with the set of NE (for example, in trust dilemmas). Therefore, in these scenarios, if RAE finds safe solutions, we would expect that the RAE solutions would overlap with the risk-dominant NE solutions. 
 

\textbf{Experiment: MDP Stag-Hunt:} We use an MDP-based adaptation of a classic trust dilemma game from \citep{prosocial} where there is a payoff-dominant equilibrium (chasing the stag) and a risk-dominant equilibrium (gathering plants). 


\textbf{Baselines} As this is a MDP task we integrate Nash, Uniform, Self-Play, THPE, QRE, described in Sec. \ref{sec:related_work} and Appendix \ref{appendix:baselines}, as the meta-distributions in the PSRO framework, denoted as e.g. PSRO-\{Nash\}. In this setting we limit our baselines to PSRO-\{Variant\} algorithms, and do not consider non-population risk-aversion algorithms (standard in PSRO literature). Full details in Appendix \ref{appendix:baselines}.

\textbf{Results} Fig. (\ref{fig:stag_hunt}B) shows all baselines arrive at the payoff-dominant stag catching strategy, whereas RAE arrives at the 'risk-dominant' plant gathering strategy. The baselines solution is risky due to its susceptibility to coordination failure i.e. when only one agent hunts the stag. This could occur frequently, for example, in situations where agents are not able to communicate with each other, or agents do not know each others strategies. For example, we show this by placing a PSRO-RAE population and a PSRO-Nash population into the environment together as co-players, shown in Fig. (\ref{fig:stag_hunt}C). The Nash population still attempts to hunt the stag (unable to communicate, has a fixed strategy), but in this case the RAE population is still gathering plants - leading to the Nash population being caught by the stag many times, and the RAE population remaining safe.
 

 
 \subsection{Are RAE strategies safe?} 
 \label{sec:exps-drive}

\textbf{Motivation} RAE is designed to avoid strategies that are overly susceptible to other agents strategies by limiting UVar. We examine how RAE acts in a larger-scale MARL autonomous driving setting where avoiding any large negative outcome (e.g. crashing) is critical, in particular in the presence of strategies that can look overly advantageous unless coordination between agents fails. 

\textbf{Experiment: Autonomous Driving Scenario \citep{highway-env}:} MDP recreation of Fig. (\ref{fig:self-drive}) scenario; two-way traffic with slow-moving vehicles and faster moving agents behind that may be interested in overtaking.  

\textbf{Baselines} MDP task, refer to Sec. \ref{sec:exps-stag}.

\textbf{Results} Results are in Fig. (\ref{fig:drive_res}). In Table (\ref{fig:drive_res}A) we provide a collection of environment metrics where the average value is based on 50 environment episodes and the standard deviation is from 5 training seeds. In terms of EU and UVar, RAE actually outperforms the baselines considerably, whilst \textit{maintaining strong worst-case performance}. Notably, RAE arrives at a strategy that very rarely crashes, and nearly always arrives at the final destination. The same conclusion can not be drawn for the baselines which often crash and fail to reach the destination. To understand this better, in Fig. (\ref{fig:drive_res}B, C) we provide position heat-maps of a PSRO-RAE and a PSRO-Nash car respectively. RAE executes the safe strategy, i.e. follow behind until all vehicles in the on-coming lane have passed and then proceeds to overtake. This strategy remains sensitive to the risk-element of the environment, overtaking and crashing, which is our desired outcome. On the other hand, the Nash strategy overtakes straight away and nearly always ends up in a crash due to congestion in the middle of the road.  

 
 \section{Conclusion}
 We introduce a new risk-averse equilibrium, RAE, based on agents considering the variance of the utility function alongside the expected value. Theoretically, we prove the existence and solvability of RAE and provide methods for arriving at an RAE in both small and large-scale game settings. Empirically, we show that our RAE is able to locate minimum variance solutions for any EU, act as a NE selection method in the presence of risk-dominant NE, and is effective at finding a safe equilibrium in a safety-sensitive autonomous driving environment. Avenues for future work should focus on the limitations of the current RAE approach, namely non-convergence guarantees in certain classes of games and the fact that RAE minimises upside and downside variance, where minimising downside variance only would be a desirable property.

\clearpage

\bibliography{example_paper}

\begin{thebibliography}{47}
\providecommand{\natexlab}[1]{#1}
\providecommand{\url}[1]{\texttt{#1}}
\expandafter\ifx\csname urlstyle\endcsname\relax
  \providecommand{\doi}[1]{doi: #1}\else
  \providecommand{\doi}{doi: \begingroup \urlstyle{rm}\Url}\fi

\bibitem[Beckmann et~al.(1956)Beckmann, McGuire, and
  Winsten]{beckmann1956studies}
Beckmann, M., McGuire, C.~B., and Winsten, C.~B.
\newblock Studies in the economics of transportation.
\newblock 1956.

\bibitem[Bielefeld(1988)]{bielefeld1988reexamination}
Bielefeld, R.~S.
\newblock Reexamination of the perfectness concept for equilibrium points in
  extensive games.
\newblock In \emph{Models of Strategic Rationality}, pp.\  1--31. Springer,
  1988.

\bibitem[Bisi et~al.(2019)Bisi, Sabbioni, Vittori, Papini, and
  Restelli]{bisi2019risk}
Bisi, L., Sabbioni, L., Vittori, E., Papini, M., and Restelli, M.
\newblock Risk-averse trust region optimization for reward-volatility
  reduction.
\newblock \emph{arXiv preprint arXiv:1912.03193}, 2019.

\bibitem[Cherukuri(2019)]{cherukuri2019sample}
Cherukuri, A.
\newblock Sample average approximation of cvar-based wardrop equilibrium in
  routing under uncertain costs.
\newblock In \emph{2019 IEEE 58th Conference on Decision and Control (CDC)},
  pp.\  3164--3169. IEEE, 2019.

\bibitem[Dinh et~al.(2021)Dinh, Yang, Tian, Nieves, Slumbers, Mguni, and
  Wang]{dinh2021online}
Dinh, L.~C., Yang, Y., Tian, Z., Nieves, N.~P., Slumbers, O., Mguni, D.~H., and
  Wang, J.
\newblock Online double oracle.
\newblock \emph{arXiv preprint arXiv:2103.07780}, 2021.

\bibitem[Eriksson et~al.(2022)Eriksson, Basu, Alibeigi, and
  Dimitrakakis]{eriksson2022risk}
Eriksson, H., Basu, D., Alibeigi, M., and Dimitrakakis, C.
\newblock Risk-sensitive bayesian games for multi-agent reinforcement learning
  under policy uncertainty.
\newblock \emph{arXiv preprint arXiv:2203.10045}, 2022.

\bibitem[Feng et~al.(2021)Feng, Slumbers, Wan, Liu, McAleer, Wen, Wang, and
  Yang]{feng2021neural}
Feng, X., Slumbers, O., Wan, Z., Liu, B., McAleer, S., Wen, Y., Wang, J., and
  Yang, Y.
\newblock Neural auto-curricula in two-player zero-sum games.
\newblock \emph{Advances in Neural Information Processing Systems}, 34, 2021.

\bibitem[Fiat \& Papadimitriou(2010)Fiat and Papadimitriou]{fiat2010players}
Fiat, A. and Papadimitriou, C.
\newblock When the players are not expectation maximizers.
\newblock In \emph{International Symposium on Algorithmic Game Theory}, pp.\
  1--14. Springer, 2010.

\bibitem[Fudenberg \& Kreps(1993)Fudenberg and Kreps]{fudenberg1993learning}
Fudenberg, D. and Kreps, D.~M.
\newblock Learning mixed equilibria.
\newblock \emph{Games and economic behavior}, 5\penalty0 (3):\penalty0
  320--367, 1993.

\bibitem[Gal \& Grosz(2022)Gal and Grosz]{gal2022multi}
Gal, K. and Grosz, B.~J.
\newblock Multi-agent systems: Technical \& ethical challenges of functioning
  in a mixed group.
\newblock \emph{Daedalus}, 151\penalty0 (2):\penalty0 114--126, 2022.

\bibitem[Ganzfried(2020)]{ganzfried2020fictitious}
Ganzfried, S.
\newblock Fictitious play outperforms counterfactual regret minimization.
\newblock \emph{arXiv preprint arXiv:2001.11165}, 2020.

\bibitem[Gao et~al.(2021)Gao, Lui, and Hernandez-Leal]{gao2021robust}
Gao, Y., Lui, K. Y.~C., and Hernandez-Leal, P.
\newblock Robust risk-sensitive reinforcement learning agents for trading
  markets.
\newblock \emph{arXiv preprint arXiv:2107.08083}, 2021.

\bibitem[Goldberg et~al.(2013)Goldberg, Savani, S{\o}rensen, and
  Ventre]{goldberg2013approximation}
Goldberg, P.~W., Savani, R., S{\o}rensen, T.~B., and Ventre, C.
\newblock On the approximation performance of fictitious play in finite games.
\newblock \emph{International Journal of Game Theory}, 42\penalty0
  (4):\penalty0 1059--1083, 2013.

\bibitem[Harsanyi et~al.(1988)Harsanyi, Selten, et~al.]{harsanyi1988general}
Harsanyi, J.~C., Selten, R., et~al.
\newblock A general theory of equilibrium selection in games.
\newblock \emph{MIT Press Books}, 1, 1988.

\bibitem[Lanctot et~al.(2017)Lanctot, Zambaldi, Gruslys, Lazaridou, Tuyls,
  P{\'e}rolat, Silver, and Graepel]{lanctot2017unified}
Lanctot, M., Zambaldi, V., Gruslys, A., Lazaridou, A., Tuyls, K., P{\'e}rolat,
  J., Silver, D., and Graepel, T.
\newblock A unified game-theoretic approach to multiagent reinforcement
  learning.
\newblock In \emph{Proceedings of the 31st International Conference on Neural
  Information Processing Systems}, pp.\  4193--4206, 2017.

\bibitem[Leurent(2018)]{highway-env}
Leurent, E.
\newblock An environment for autonomous driving decision-making.
\newblock \url{https://github.com/eleurent/highway-env}, 2018.

\bibitem[Lianeas et~al.(2019)Lianeas, Nikolova, and
  Stier-Moses]{lianeas2019risk}
Lianeas, T., Nikolova, E., and Stier-Moses, N.~E.
\newblock Risk-averse selfish routing.
\newblock \emph{Mathematics of Operations Research}, 44\penalty0 (1):\penalty0
  38--57, 2019.

\bibitem[Markowitz(1991)]{markowitz1991foundations}
Markowitz, H.~M.
\newblock Foundations of portfolio theory.
\newblock \emph{The journal of finance}, 46\penalty0 (2):\penalty0 469--477,
  1991.

\bibitem[McAleer et~al.(2020)McAleer, Lanier, Fox, and
  Baldi]{mcaleer2020pipeline}
McAleer, S., Lanier, J., Fox, R., and Baldi, P.
\newblock Pipeline {PSRO}: A scalable approach for finding approximate nash
  equilibria in large games.
\newblock In \emph{Advances in Neural Information Processing Systems
  (NeurIPS)}, 2020.

\bibitem[McAleer et~al.(2021)McAleer, Lanier, Wang, Baldi, and
  Fox]{mcaleer2021xdo}
McAleer, S., Lanier, J., Wang, K., Baldi, P., and Fox, R.
\newblock {XDO}: A double oracle algorithm for extensive-form games.
\newblock \emph{Advances in Neural Information Processing Systems (NeurIPS)},
  2021.

\bibitem[McAleer et~al.(2022{\natexlab{a}})McAleer, Farina, Lanctot, and
  Sandholm]{mcaleer2022escher}
McAleer, S., Farina, G., Lanctot, M., and Sandholm, T.
\newblock Escher: Eschewing importance sampling in games by computing a history
  value function to estimate regret.
\newblock \emph{arXiv preprint arXiv:2206.04122}, 2022{\natexlab{a}}.

\bibitem[McAleer et~al.(2022{\natexlab{b}})McAleer, Lanier, Wang, Baldi, Fox,
  and Sandholm]{mcaleer2022self}
McAleer, S., Lanier, J., Wang, K., Baldi, P., Fox, R., and Sandholm, T.
\newblock Self-play psro: Toward optimal populations in two-player zero-sum
  games.
\newblock \emph{arXiv preprint arXiv:2207.06541}, 2022{\natexlab{b}}.

\bibitem[McAleer et~al.(2022{\natexlab{c}})McAleer, Wang, Lanctot, Lanier,
  Baldi, and Fox]{mcaleer2022anytime}
McAleer, S., Wang, K., Lanctot, M., Lanier, J., Baldi, P., and Fox, R.
\newblock Anytime optimal psro for two-player zero-sum games.
\newblock \emph{arXiv preprint arXiv:2201.07700}, 2022{\natexlab{c}}.

\bibitem[McKelvey \& Palfrey(1995)McKelvey and Palfrey]{mckelvey1995quantal}
McKelvey, R.~D. and Palfrey, T.~R.
\newblock Quantal response equilibria for normal form games.
\newblock \emph{Games and economic behavior}, 10\penalty0 (1):\penalty0 6--38,
  1995.

\bibitem[McMahan et~al.(2003)McMahan, Gordon, and Blum]{mcmahan2003planning}
McMahan, H.~B., Gordon, G.~J., and Blum, A.
\newblock Planning in the presence of cost functions controlled by an
  adversary.
\newblock In \emph{Proceedings of the 20th International Conference on Machine
  Learning (ICML-03)}, pp.\  536--543, 2003.

\bibitem[Meir \& Parkes(2015)Meir and Parkes]{meir2015congestion}
Meir, R. and Parkes, D.
\newblock Congestion games with distance-based strict uncertainty.
\newblock In \emph{Proceedings of the AAAI Conference on Artificial
  Intelligence}, volume~29, 2015.

\bibitem[Merton(1972)]{merton_1972}
Merton, R.~C.
\newblock An analytic derivation of the efficient portfolio frontier.
\newblock \emph{Journal of Financial and Quantitative Analysis}, 7\penalty0
  (4):\penalty0 1851–1872, 1972.
\newblock \doi{10.2307/2329621}.

\bibitem[Mnih et~al.(2015)Mnih, Kavukcuoglu, Silver, Rusu, Veness, Bellemare,
  Graves, Riedmiller, Fidjeland, Ostrovski, et~al.]{mnih2015human}
Mnih, V., Kavukcuoglu, K., Silver, D., Rusu, A.~A., Veness, J., Bellemare,
  M.~G., Graves, A., Riedmiller, M., Fidjeland, A.~K., Ostrovski, G., et~al.
\newblock Human-level control through deep reinforcement learning.
\newblock \emph{nature}, 518\penalty0 (7540):\penalty0 529--533, 2015.

\bibitem[Monderer \& Shapley(1996{\natexlab{a}})Monderer and
  Shapley]{monderer1996fictitious}
Monderer, D. and Shapley, L.~S.
\newblock Fictitious play property for games with identical interests.
\newblock \emph{Journal of economic theory}, 68\penalty0 (1):\penalty0
  258--265, 1996{\natexlab{a}}.

\bibitem[Monderer \& Shapley(1996{\natexlab{b}})Monderer and
  Shapley]{monderer1996potential}
Monderer, D. and Shapley, L.~S.
\newblock Potential games.
\newblock \emph{Games and economic behavior}, 14\penalty0 (1):\penalty0
  124--143, 1996{\natexlab{b}}.

\bibitem[Nash(1951)]{nash1951non}
Nash, J.
\newblock Non-cooperative games.
\newblock \emph{Annals of mathematics}, pp.\  286--295, 1951.

\bibitem[Nikolova \& Stier-Moses(2012)Nikolova and
  Stier-Moses]{nikolova2012stochastic}
Nikolova, E. and Stier-Moses, N.
\newblock Stochastic selfish routing.
\newblock \emph{ACM SIGecom Exchanges}, 11\penalty0 (1):\penalty0 21--25, 2012.

\bibitem[Nikolova \& Stier-Moses(2014)Nikolova and
  Stier-Moses]{nikolova2014mean}
Nikolova, E. and Stier-Moses, N.~E.
\newblock A mean-risk model for the traffic assignment problem with stochastic
  travel times.
\newblock \emph{Operations Research}, 62\penalty0 (2):\penalty0 366--382, 2014.

\bibitem[Ord{\'o}{\~n}ez \& Stier-Moses(2010)Ord{\'o}{\~n}ez and
  Stier-Moses]{ordonez2010wardrop}
Ord{\'o}{\~n}ez, F. and Stier-Moses, N.~E.
\newblock Wardrop equilibria with risk-averse users.
\newblock \emph{Transportation Science}, 44\penalty0 (1):\penalty0 63--86,
  2010.

\bibitem[Perez-Nieves et~al.(2021)Perez-Nieves, Yang, Slumbers, Mguni, Wen, and
  Wang]{pmlr-v139-perez-nieves21a}
Perez-Nieves, N., Yang, Y., Slumbers, O., Mguni, D.~H., Wen, Y., and Wang, J.
\newblock Modelling behavioural diversity for learning in open-ended games.
\newblock In Meila, M. and Zhang, T. (eds.), \emph{Proceedings of the 38th
  International Conference on Machine Learning}, volume 139 of
  \emph{Proceedings of Machine Learning Research}, pp.\  8514--8524. PMLR,
  18--24 Jul 2021.
\newblock URL \url{https://proceedings.mlr.press/v139/perez-nieves21a.html}.

\bibitem[Perolat et~al.(2022)Perolat, De~Vylder, Hennes, Tarassov, Strub,
  de~Boer, Muller, Connor, Burch, Anthony, et~al.]{perolat2022mastering}
Perolat, J., De~Vylder, B., Hennes, D., Tarassov, E., Strub, F., de~Boer, V.,
  Muller, P., Connor, J.~T., Burch, N., Anthony, T., et~al.
\newblock Mastering the game of stratego with model-free multiagent
  reinforcement learning.
\newblock \emph{Science}, 378\penalty0 (6623):\penalty0 990--996, 2022.

\bibitem[Peysakhovich \& Lerer(2017)Peysakhovich and Lerer]{prosocial}
Peysakhovich, A. and Lerer, A.
\newblock Prosocial learning agents solve generalized stag hunts better than
  selfish ones, 2017.
\newblock URL \url{https://arxiv.org/abs/1709.02865}.

\bibitem[Piliouras et~al.(2016)Piliouras, Nikolova, and
  Shamma]{piliouras2016risk}
Piliouras, G., Nikolova, E., and Shamma, J.~S.
\newblock Risk sensitivity of price of anarchy under uncertainty.
\newblock \emph{ACM Transactions on Economics and Computation (TEAC)},
  5\penalty0 (1):\penalty0 1--27, 2016.

\bibitem[Qiu et~al.(2021)Qiu, Wang, Yu, Wang, He, An, Obraztsova, and
  Rabinovich]{qiu2021rmix}
Qiu, W., Wang, X., Yu, R., Wang, R., He, X., An, B., Obraztsova, S., and
  Rabinovich, Z.
\newblock Rmix: Learning risk-sensitive policies for cooperative reinforcement
  learning agents.
\newblock \emph{Advances in Neural Information Processing Systems},
  34:\penalty0 23049--23062, 2021.

\bibitem[Raffin et~al.(2021)Raffin, Hill, Gleave, Kanervisto, Ernestus, and
  Dormann]{stable-baselines3}
Raffin, A., Hill, A., Gleave, A., Kanervisto, A., Ernestus, M., and Dormann, N.
\newblock Stable-baselines3: Reliable reinforcement learning implementations.
\newblock \emph{Journal of Machine Learning Research}, 22\penalty0
  (268):\penalty0 1--8, 2021.
\newblock URL \url{http://jmlr.org/papers/v22/20-1364.html}.

\bibitem[Robinson(1951)]{10.2307/1969530}
Robinson, J.
\newblock An iterative method of solving a game.
\newblock \emph{Annals of Mathematics}, 54\penalty0 (2):\penalty0 296--301,
  1951.
\newblock ISSN 0003486X.
\newblock URL \url{http://www.jstor.org/stable/1969530}.

\bibitem[Royset(2022)]{royset2022risk}
Royset, J.~O.
\newblock Risk-adaptive approaches to learning and decision making: A survey.
\newblock \emph{arXiv preprint arXiv:2212.00856}, 2022.

\bibitem[Sokota et~al.(2022)Sokota, D'Orazio, Kolter, Loizou, Lanctot,
  Mitliagkas, Brown, and Kroer]{sokota2022unified}
Sokota, S., D'Orazio, R., Kolter, J.~Z., Loizou, N., Lanctot, M., Mitliagkas,
  I., Brown, N., and Kroer, C.
\newblock A unified approach to reinforcement learning, quantal response
  equilibria, and two-player zero-sum games.
\newblock \emph{arXiv preprint arXiv:2206.05825}, 2022.

\bibitem[Wardrop(1952)]{wardrop1952road}
Wardrop, J.~G.
\newblock Road paper. some theoretical aspects of road traffic research.
\newblock \emph{Proceedings of the institution of civil engineers}, 1\penalty0
  (3):\penalty0 325--362, 1952.

\bibitem[Wellman(2006)]{wellman2006methods}
Wellman, M.~P.
\newblock Methods for empirical game-theoretic analysis.
\newblock In \emph{AAAI}, pp.\  1552--1556, 2006.

\bibitem[Yekkehkhany et~al.(2020)Yekkehkhany, Murray, and
  Nagi]{yekkehkhany2020riskaverse}
Yekkehkhany, A., Murray, T., and Nagi, R.
\newblock Risk-averse equilibrium for games, 2020.

\bibitem[Zhang et~al.(2020)Zhang, Liu, and Whiteson]{zhang2020mean}
Zhang, S., Liu, B., and Whiteson, S.
\newblock Mean-variance policy iteration for risk-averse reinforcement
  learning.
\newblock \emph{arXiv preprint arXiv:2004.10888}, 2020.

\end{thebibliography}
\bibliographystyle{icml2023}

\newpage
\appendix
\onecolumn

\section{Full Proofs}
\label{appendix:proofs}
\subsection{Proposition \ref{prop:min-var-solution} [Minimum Variance Solution]}
\label{appendix:proofs-min-var-solution}

The solution to the Eq. \ref{eq:br-map} provides the same solutions to the following:,
\begin{equation}
\begin{aligned}
\boldsymbol{\sigma}^* \in \argmin_{\boldsymbol{\sigma}}  \boldsymbol{\sigma}^{T} \cdot \Sigma_{\tM} \cdot \boldsymbol{\varsigma} \\
\text{s.t. } \boldsymbol{\sigma}^{T} \cdot \tM \cdot \boldsymbol{\sigma} \geq \mu_{\text{b}}\\
\sigma(a) \geq 0 \text{  } \forall a \in A\\
\boldsymbol{\sigma}^{T}\textbf{1} = 1
\end{aligned}
\end{equation}
where $\mu_{\text{b}} \in \mathbb{R}$ is the lowest level of expected return that the actor is willing to accept.

\begin{proof}
\citep{merton_1972} shows by a Lagrange multiplier argument that the optimisation problem, 
\begin{equation}
\begin{aligned}
\boldsymbol{\sigma}^* \in \argmin_{\boldsymbol{\sigma}}  \boldsymbol{\sigma}^{T} \cdot \Sigma_{\tM} \cdot \boldsymbol{\sigma} \\
\text{s.t. } \boldsymbol{\sigma}^{T} \cdot \tM \cdot \boldsymbol{\varsigma} \geq \mu_{\text{b}}\\
\sigma(a) \geq 0 \text{  } \forall a \in A\\
\boldsymbol{\sigma}^{T}\textbf{1} = 1
\end{aligned}
\end{equation}

can be rewritten as 
\begin{equation}
\begin{aligned}
\boldsymbol{\sigma}^* \in \argmin_{\boldsymbol{\sigma}}  \boldsymbol{\sigma}^{T} \cdot \Sigma_{\tM} \cdot \boldsymbol{\sigma} - \tau\Big(\boldsymbol{\sigma}^{T} \cdot \tM \cdot \boldsymbol{\varsigma}\Big)\\
\text{s.t. } 
\sigma(a) \geq 0 \text{  } \forall a \in A\\
\boldsymbol{\sigma}^{T}\textbf{1} = 1
\end{aligned}
\end{equation}

which can be equivalently expressed as,
\begin{equation}
\begin{aligned}
\boldsymbol{\sigma}^* \in \argmin_{\boldsymbol{\sigma}}  -\Big(\boldsymbol{\sigma}^{T} \cdot \tM \cdot \boldsymbol{\varsigma} - \lambda \boldsymbol{\sigma}^{T} \cdot \Sigma_{\tM} \cdot \boldsymbol{\sigma}\Big)\\
\text{s.t. } 
\sigma(a) \geq 0 \text{  } \forall a \in A\\
\boldsymbol{\sigma}^{T}\textbf{1} = 1
\end{aligned}
\end{equation}

where $\lambda = \frac{1}{\tau}$. 

\end{proof}

\subsection{Theorem \ref{thrm:existence} [RAE Existence]}
\label{appendix:proofs-existence}

For any finite N-player game where each player $i$ has a finite $k$ number of pure strategies, $A^i = \{a^i_1, ..., a^i_k\}$, an RAE exists

\begin{proof}
We base our proof on Kakutani's Fixed Point Theorem

\begin{addmargin}[2.5em]{2.5em}
\begin{lemma*}[Kakutani Fixed Point Theorem] Let $A$ be a non-empty subset of a finite dimensional Euclidean space. Let $f: A \rightrightarrows A$ be a correspondence, with $x \in A \longmapsto f(x) \subseteq A$, satisfying the following conditions:
\begin{enumerate}
    \item $A$ is a compact and convex set.
    \item $f(x)$ is non-empty for all $x \in A$.
    \item $f(x)$ is a convex-valued correspondence: for all $x \in A$, $f(x)$ is a convex set.
    \item $f(x)$ has a closed graph: that is, if $\{x^n, y^n\} \rightarrow \{x,y\}$ with $y^n \in f(x^n)$, then $y \in f(x)$.
\end{enumerate}

Then, $f$ has a fixed point, that is, there exists some $x \in A$, such that $x \in f(x)$.
\end{lemma*}
\end{addmargin}

We define our best-response function as $B_i(\boldsymbol{\sigma}_{-i}) = \argmax_{a \in \Delta_i} r^i(a, \boldsymbol{\sigma}_{-i})$ where $u_i$ is defined as in Eq. (5) and by definition $s$ must satisfy all of the properties of a proper mixed-strategy, and the best-response correspondence is $B: \Delta \rightrightarrows \Delta$ such that for all $\boldsymbol{\sigma} \in \Delta$, we have:

\begin{equation}
    B(\boldsymbol{\sigma}) = [B_i(\boldsymbol{\sigma}_{-i})]_{i\in N}
\end{equation}

We show that $B(\boldsymbol{\sigma})$ satisfies the conditions of Kakutani's Fixed Point Theorem

\begin{enumerate}
    \item \textit{$\Delta$ is compact, convex and non-empty.}
    
    By definition 
    \begin{equation}
        \Delta = \Pi_{i \in N} \Delta_i
    \end{equation}
    where each $\Delta_i = \{a | \sum_j a_j = 1 \}$ is a simplex of dimension $|A^i| - 1$, thus each $\Delta_i$ is closed and bounded, and thus compact. Their product set is also compact. 
    
    \item \textit{$B(\boldsymbol{\sigma})$ is non-empty.} 
    
    By the definition of $B_i(\boldsymbol{\sigma}_{-i})$ where $\Delta_i$ is non-empty and compact, and $r^i$ is a quadratic and hence a polynomial function in $a$. It is known that all polynomial functions are continuous, we can invoke Weirstrass's Extreme Value Theorem which states
    
    \begin{lemma*}
    If a real valued-function $f$ is continuous on the closed interval $[a,b]$, then $f$ must attain a maximum and a minimum, each at least once. That is, there exist numbers $c$ and $d$ in $[a,b]$ such that:
    \begin{equation*}
        f(c) \geq f(x) \geq f(d) \quad \forall x \in [a,b]
    \end{equation*}
    \end{lemma*}
    
    Therefore, as $\Delta_i$ is non-empty and compact and $r^i$ is continuous in $a$, $B_i(\boldsymbol{\sigma}_{-i})$ is non-empty, and therefore $B(\boldsymbol{\sigma})$ is also non-empty.
    
    \item \textit{$B(\boldsymbol{\sigma})$ is a convex-valued correspondence}.
    
    Equivalently, $B(\boldsymbol{\sigma}) \subset \Delta$ is convex if and only if $B_i(\boldsymbol{\sigma}_{-i})$ is convex for all $i$. 
    
    In order to show that $B_i(\boldsymbol{\sigma}_{-i}$) is convex for all $i$, we instead show that the Quadratic Programme defined by Eq. (6) is a special case of convex optimisation under certain conditions, and therefore by definition has a feasible set which is a convex set. 
    
    A \textit{convex optimisation problem} is one of the form,
    
    \begin{equation} 
    \begin{aligned}
    \text{minimize} \quad &f_0(x)\\
    \text{s.t. } &f_i(x) < 0, \text{  } i=1,...,m\\
    &a_i^{T}x = b_i, \quad i=1,...,p
    \end{aligned}
    \end{equation}
    
    where $f_0, ..., f_m$ are convex functions. The requirements for a problem to be a convex optimisation problem are:
    
    \begin{enumerate}
        \item the objective function must be convex
        \item the inequality constraint functions must be convex
        \item the equality constraint functions $h_i(x)=a_i^{T}x = b_i$ must be affine
    \end{enumerate}
    
    We note that a quadratic form $\mathbf{x}^{T} \boldsymbol{A} \mathbf{x}$ is convex if $\boldsymbol{A}$ is positive semi-definite, and strictly convex if $\boldsymbol{A}$ is positive definite. In our constrained optimisation, the quadratic term $\boldsymbol{\sigma}^{T} \Sigma \boldsymbol{\sigma}$ is always guaranteed to be at least convex as $\Sigma$, the covariance matrix, is always at least PSD. Therefore, our objective function is convex. Additionally, it is easy to see that our inequality constraint functions are also convex and that our equality constraint function is affine. Therefore, our Quadratic Programme is an instance of a convex optimisation problem.
    
    Importantly, the feasible set of a convex optimisation problem is convex, since it is the intersection of the domain of the problem
    
    \begin{equation}
        \mathcal{D} = \bigcap_{i=0}^m \textbf{dom} f_i,
    \end{equation}
    , which itself is a convex set. 
    
    Therefore, for all members of the feasible set $x, y \in B_i(\boldsymbol{\sigma}_{-i})$ and all $\theta\in [0,1]$ we have that $\theta x + (1-\theta)y \in S$ and we have a convex-valued correspondence.
    
    \item \textit{$B(\boldsymbol{\sigma})$ has a closed graph.}
    
    Suppose to obtain a contradiction, that $B(\boldsymbol{\sigma})$ does not have a closed graph. Then, there exists a sequence ($\boldsymbol{\sigma}^n, \hat{\boldsymbol{\sigma}}^n) \rightarrow (\boldsymbol{\sigma}, \hat{\boldsymbol{\sigma}})$ with $\hat{\boldsymbol{\sigma}}^n \in B(\boldsymbol{\sigma}^n)$, but $\hat{\boldsymbol{\sigma}} \notin B(\boldsymbol{\sigma})$, i.e. there exists some $i$ such that $\hat{\boldsymbol{\sigma}}_i \notin B_i(\boldsymbol{\sigma}_{-i})$. This implies that there exists some $\boldsymbol{\sigma}_i^{\prime} \in \Delta_i$ and some $\epsilon > 0$ such that 
    
    \begin{equation}
        r_i(\boldsymbol{\sigma}_i^{\prime}, \boldsymbol{\sigma}_{-i}) > r_i(\hat{\boldsymbol{\sigma}}_i, \boldsymbol{\sigma}_{-i}) + 3\epsilon.
    \end{equation}
    
    By the continuity of $r_i$ and the fact that $\boldsymbol{\sigma}_{-i}^n \rightarrow \boldsymbol{\sigma}_{-i}$, we have for sufficiently large $n$,
    
    \begin{equation}
        r_i(\boldsymbol{\sigma}_i^{\prime}, \boldsymbol{\sigma}_{-i}^n) \geq r_i(\boldsymbol{\sigma}_i^{\prime}, \boldsymbol{\sigma}_{-i}) - \epsilon.
    \end{equation}
    
    and combining the preceding two relations we obtain
    
    \begin{equation}
        r_i(\boldsymbol{\sigma}_i^{\prime}, \boldsymbol{\sigma}_{-i}^n) > r_i(\hat{\boldsymbol{\sigma}}_i, \boldsymbol{\sigma}_{-i}) + 2\epsilon \geq r_i(\hat{\boldsymbol{\sigma}}_i^n, \boldsymbol{\sigma}_{-i}^n) + \epsilon
    \end{equation}
    
    where the second relation follows from the continuity of $r_i$. This contradicts the assumption that $\hat{\boldsymbol{\sigma}}_i^n \in B(\boldsymbol{\sigma}_{-i}^n)$ and completes the proof.
\end{enumerate}

Therefore, $B(\boldsymbol{\sigma})$ satisfies the conditions of Kakutani's Fixed Point Theorem, and therefore if $\boldsymbol{\sigma}^* \in B(\boldsymbol{\sigma}^*)$ then $\boldsymbol{\sigma}^*$ is an equilibrium.
\end{proof}

\subsection{Proposition \ref{prop:sfp-convergence} [SFP Convergence]}

Replacing the best-response Eq. (\ref{eq:sfp-br}) with the best-response map Eq. (\ref{eq:br-map}) satisfies the conditions of $\bar{u}^i$ for a SFP process. 

\begin{proof}
For the perturbed utility function $\bar{u}^i$ to be permissible in an SFP process, there exist two conditions:

\begin{enumerate}
    \item That there exists a unique global solution to $\bar{u}^i$.
    \item That the $\argmax$ assigns strictly positive probability to all pure strategies.
\end{enumerate}

We let $\bar{u}^i$ be replaced by the best-response map Eq. \ref{eq:br-map} and show that the 2 conditions noted above are met.

\textbf{Condition 1 -} To show that there exists a global solution to $\bar{u}^i$ we need to show that $\bar{u}^i$ is strictly concave which guarantees a unique global maximum. 

As the EU term $u^i(\boldsymbol{\sigma}, Z_t^{-i} \tM)$ is linear, we therefore require that the perturbation term $v^i(\boldsymbol{\sigma})$ is strictly convex such that the perturbed utility function $\bar{u}^i = \argmax_{\boldsymbol{\sigma}} u^i(\boldsymbol{\sigma}, Z_t^{-i}, \tM) - \lambda v^i(\boldsymbol{\sigma})$ is strictly concave. In \ref{appendix:proofs-existence} we have already shown that, as long as the covariance matrix $\boldsymbol{\Sigma}_{\tM}$ is positive definite, then the quadratic term $\boldsymbol{\sigma}^{T} \cdot \boldsymbol{\Sigma}_{\tM} \cdot \boldsymbol{\sigma}$ is strictly convex. By design, $\boldsymbol{\Sigma}_{\tM}$ is strictly convex and therefore $\bar{u}^i$ is also strictly concave and therefore has a unique global maximum.

\textbf{Condition 2 -}
By design the QP that solves Eq. \ref{eq:br-map} is constrained such that all pure strategies receive strictly positive probability (this is by design to induce mistakes in agents). Therefore, the output of the best-response map is always within $\operatorname{int}(\Delta^i)$ and condition 2 is satisfied.
We show that our utility measure can be embedded as a version of stochastic fictitious play and therefore can be used to find equilibrium in two-player zero-sum games and potential games.

\end{proof}

\subsection{Proposition \ref{prop:sfp} [SFP is RAE]}

Suppose the SFP  sequence $\{Z_t\}$ converges to $\boldsymbol{\sigma}$ in the observed strategy sense \footnote{Convergence in the time-average $Z_t$ does not imply convergence in the actual strategy taken at each $t$, but may for example imply cyclic actual behaviour that results in average behaviour converging.}, then $\boldsymbol{\sigma}$ is a Risk-Averse equilibrium.

\begin{proof}
Assume the observed strategy has converged to $\boldsymbol{\sigma} = (\boldsymbol{\sigma}^1, \boldsymbol{\sigma}^2)$ and that the strategy is not an RAE. This implies there exists some $\boldsymbol{\sigma}^{i, \prime}$ such that:

\begin{equation}
    r^i(\boldsymbol{\sigma}^{i, \prime}, \boldsymbol{\sigma}^{-i}) > r^i(\boldsymbol{\sigma}^{i}, \boldsymbol{\sigma}^{-i})
\end{equation}

However, because $\boldsymbol{\sigma}$ has converged then the SFP sequence $\{Z_t\}$ will also converge such that $\lim_{t \rightarrow \infty} Z_t = \boldsymbol{\sigma}$ and because we are in an SFP process it must be the case that:

\begin{equation}
    r^i(\boldsymbol{\sigma}^{i}, \boldsymbol{\sigma}^{-i}) > r^i(\boldsymbol{\sigma}^{i, \prime}, \boldsymbol{\sigma}^{-i}) \quad \forall \boldsymbol{\sigma}^{i, \prime} \in \Delta^i
\end{equation}

and therefore $\boldsymbol{\sigma}^{i, \prime}$ can not be a best response to $\boldsymbol{\sigma}^{-i}$.

\end{proof}
\clearpage

\section{SFP Robustness}
\label{appendix:robust}
\begin{figure*}[h]
\centering
\includegraphics[width=0.3\linewidth]{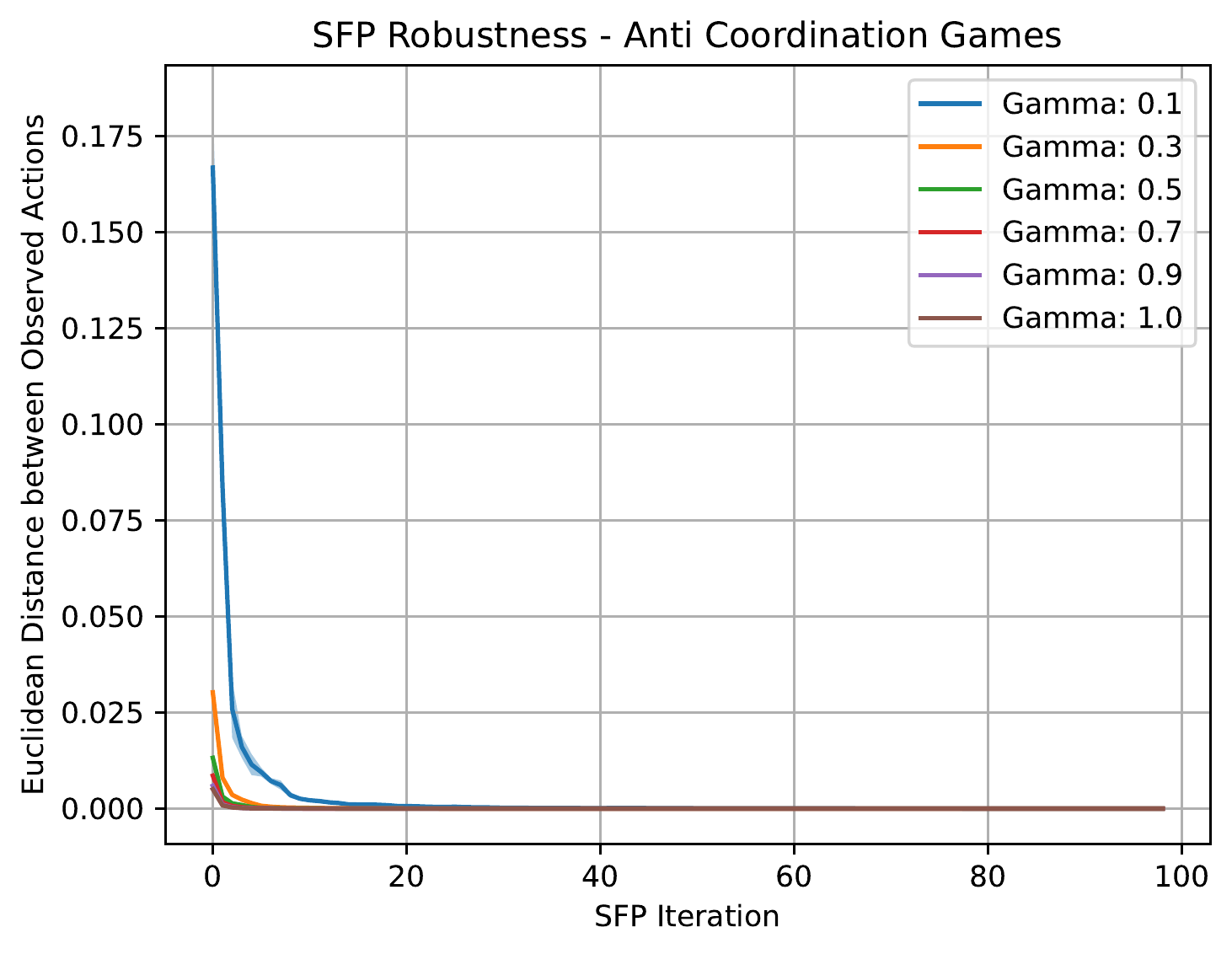}
\caption{Euclidean distance between observed actions after each iteration on randomly generated anti-coordination games. A distance of 0 implies that the process has converged.}
\end{figure*}
\vspace{-10pt}

\begin{figure*}[h]
\centering
\includegraphics[width=0.3\linewidth]{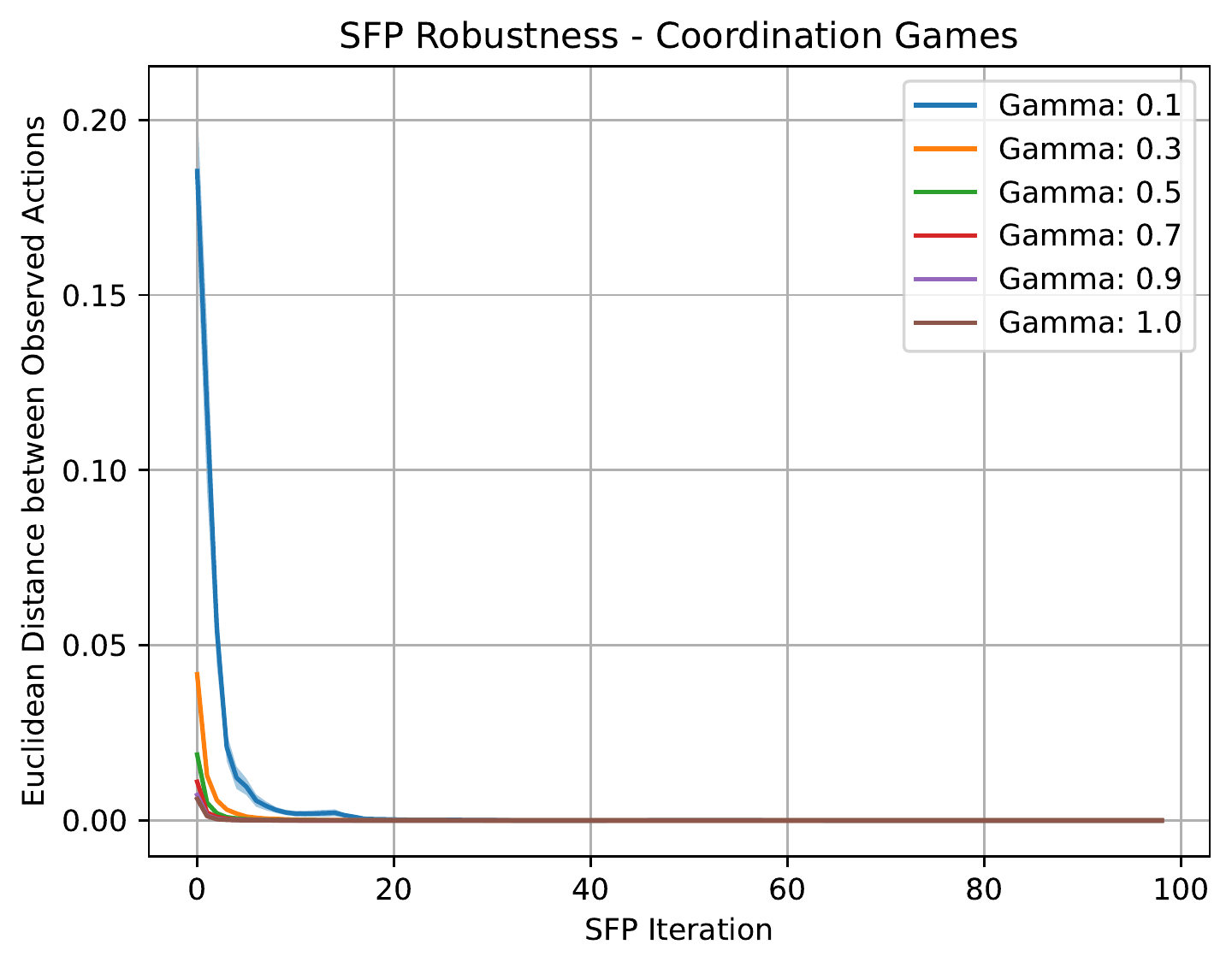}
\caption{Euclidean distance between observed actions after each iteration on randomly generated coordination games. A distance of 0 implies that the process has converged.}
\end{figure*}
\vspace{-10pt}

\begin{figure*}[h]
\centering
\includegraphics[width=0.3\linewidth]{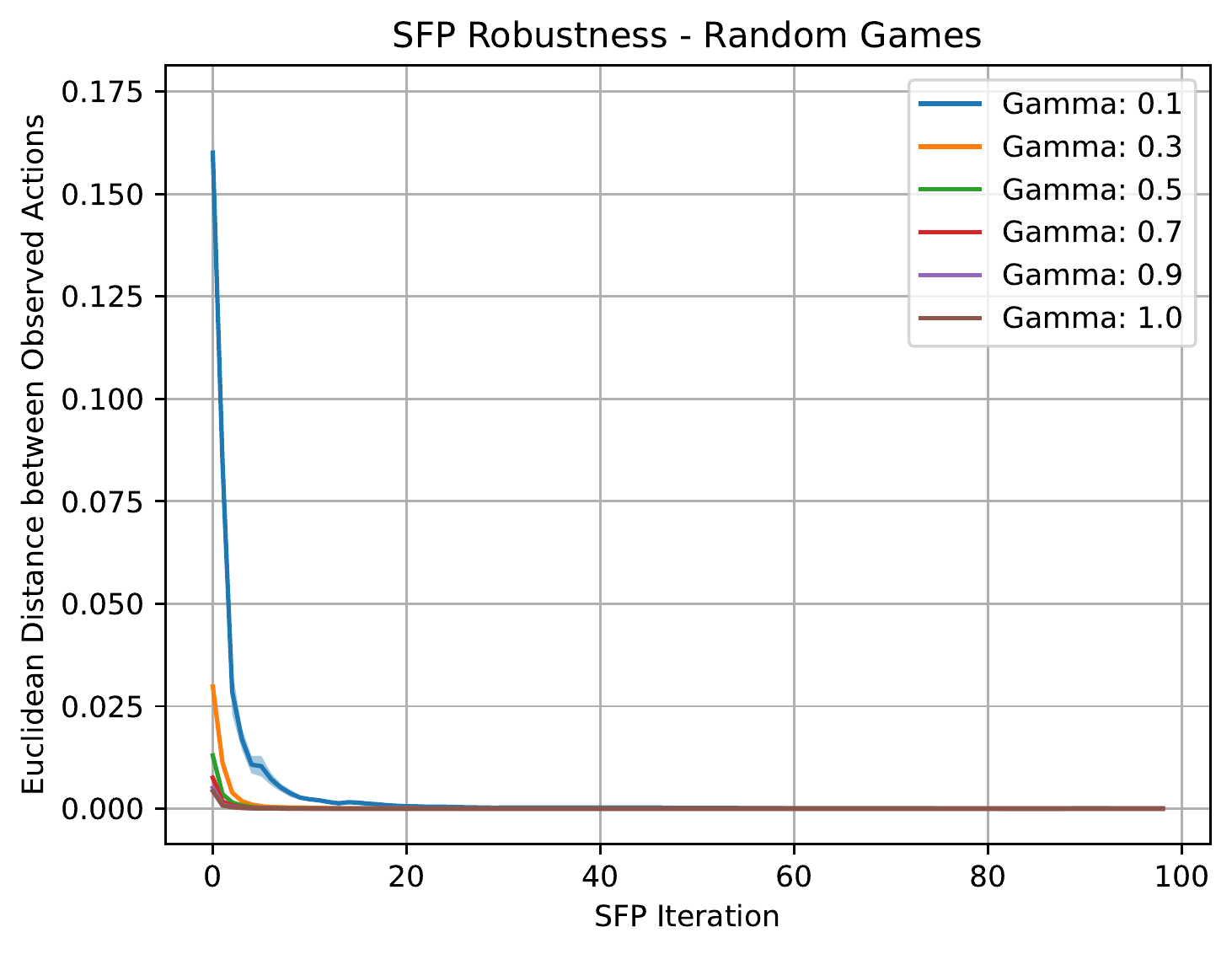}
\caption{Euclidean distance between observed actions after each iteration on randomly generated games. A distance of 0 implies that the process has converged.}
\end{figure*}

\clearpage

\section{Figure 3 Training Curves}

\begin{figure*}[h]
\centering
\includegraphics[width=0.6\linewidth]{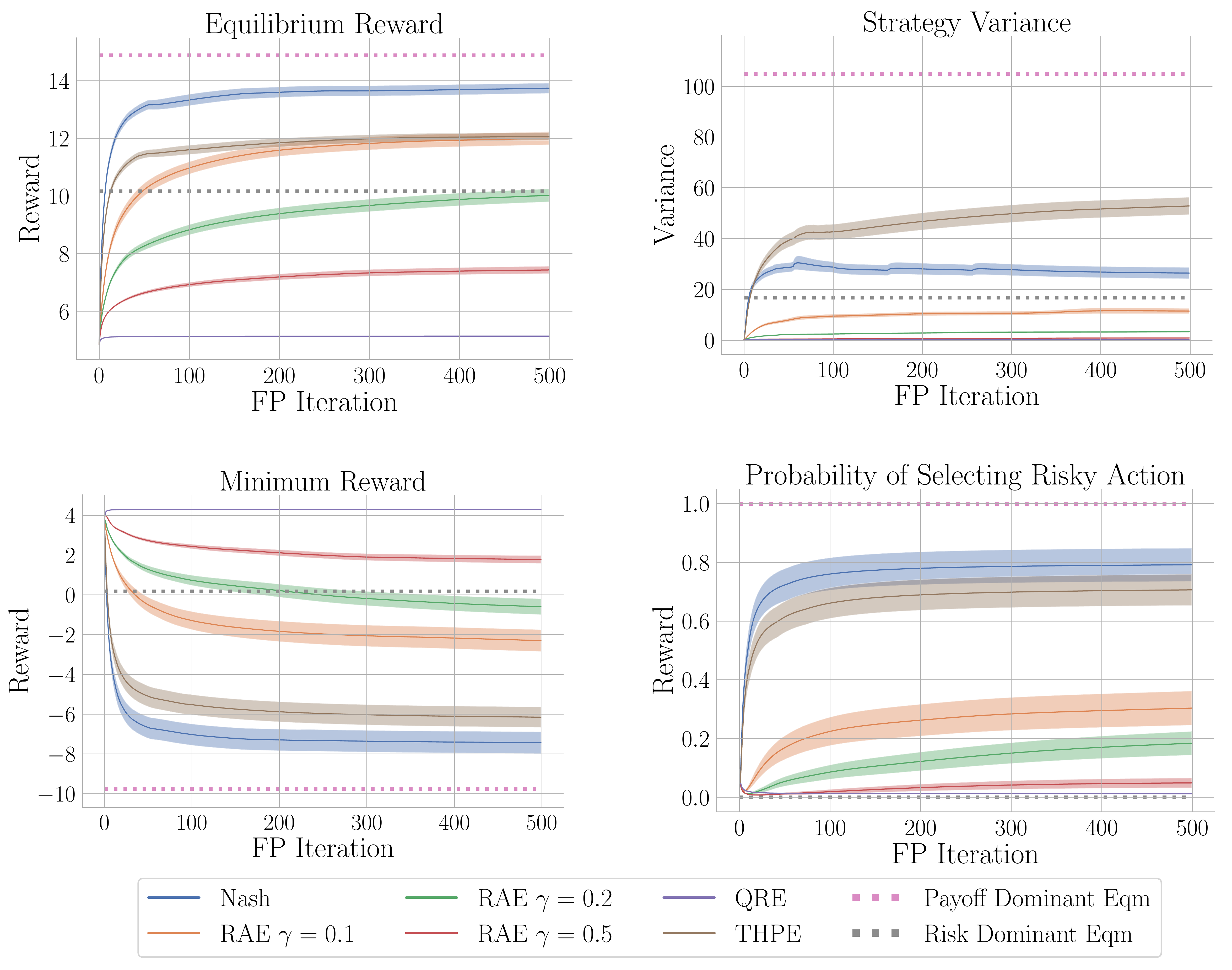}
\caption{Training curves over multiple seeds for Figure 3.}
\end{figure*}

\begin{figure*}[h]
\centering
\includegraphics[width=0.6\linewidth]{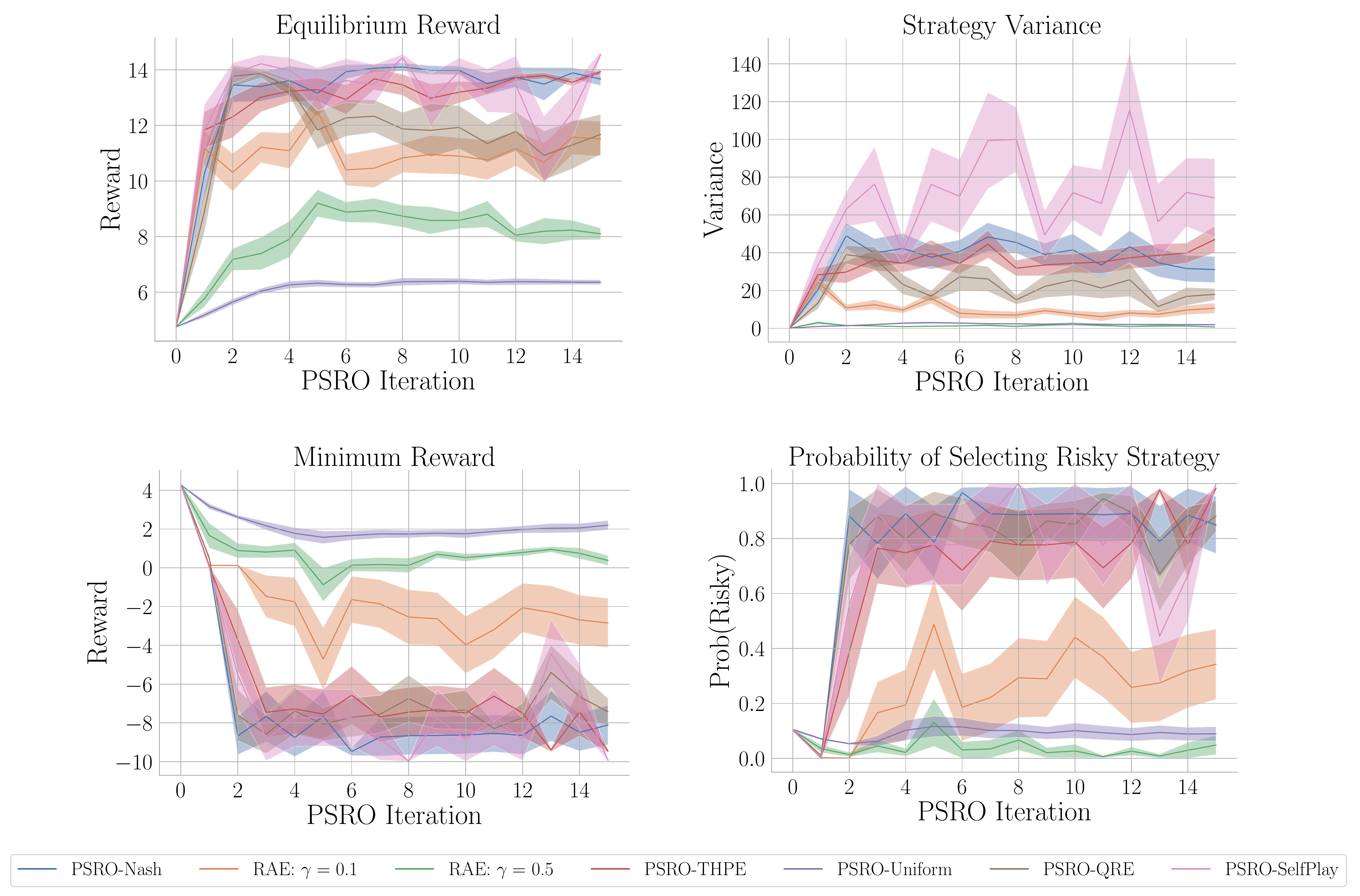}
\caption{Training curves over multiple seeds for Figure 3.}
\end{figure*}

\clearpage

\section{Pseudo-code}
\label{appendix:pseudo-code}

 \begin{algorithm}[ht]
 \caption{SFP}\label{sfp_alg}
 \begin{algorithmic}[1]
 \STATE \textbf{Initialise:} Payoff Matrix $\tM$, uniform initial time-average strategy $Z_0$.
 \STATE \textbf{for} iteration $t \in \{1,2, ... \}$ \textbf{do}:
\STATE \hspace{1em} Find best-response $\boldsymbol{\sigma}_t$ to $Z_t$ via Eq. \ref{eq:br-map}.
\STATE \hspace{1em} Update time-average strategy $Z_t$ with respect to $\boldsymbol{\sigma}_t$.
\STATE \textbf{Return:} Final time-average strategy $Z_T$. 
\end{algorithmic}
 \end{algorithm}

\begin{algorithm}[ht]
 \caption{PSRO-RAE}\label{psro_alg}
 \begin{algorithmic}[1]
 \STATE \textbf{Initialise:} the policy set $\boldsymbol{\Phi}=\prod_{i \in \mathcal{N}}\boldsymbol{\Phi}^i$, meta-game $\tM_0$, co-variance matrix $\boldsymbol{\Sigma}_{\tM_T}$
 \STATE \textbf{for} iteration $t \in \{1,2, ... \}$ \textbf{do}:
  \STATE  \hspace{1em}  \textbf{for} each player $i \in \mathcal{N}$ \textbf{do}:
\STATE \hspace{2em}  Compute meta-policy $\boldsymbol{\sigma}_t$ by SFP (Alg. \ref{sfp_alg}).
\STATE \hspace{2em} Find new  policy by Oracle: $\phi^i_t = \mathcal{O}^i(\boldsymbol{\sigma}_t)$.
\STATE \hspace{2em} Expand $\boldsymbol{\Phi}_{t+1}^i  \leftarrow \boldsymbol{\Phi}_t^i \cup \{\phi_t^i\}$.
\STATE \hspace{2em} Update meta-payoff $\tM_{t+1}$, co-variance matrix $\boldsymbol{\Sigma}_{\tM_{t+1}}$.
\STATE \textbf{Return:} $\boldsymbol{\sigma}_T$ and $\boldsymbol{\Phi}_T$. 
\end{algorithmic}
 \end{algorithm}
 
 \clearpage

\section{Hyperparameter Settings}
\label{appendix:hyperparams}


\begin{table}[ht]
\caption{Hyper-parameter settings for our experiments.}
\label{tb:hyper_baselines}
\centering
\begin{sc}
\resizebox{1. \textwidth}{!}{
\begin{tabular}{lcl}
\toprule
\textbf{Settings }& \textbf{Value} & \textbf{Description}  \\ \hline \hline
\multicolumn{3}{c}{\textbf{SFP Coordination Games}}  \\  \hline \hline
Action Dimension & 100 & Number of pure strategies available  \\
FP Iterations & 100 & Number of FP belief updates  \\
Tremble Probability & 0.001 & Probability of trembling to another strategy \\
Quantal Type & Softmax & Type of Quantal response equilibrium\\
\# of seeds & 50 & \# trials \\
\hline \hline

\multicolumn{3}{c}{\textbf{PSRO NFG Coordination Games}}  \\  \hline \hline
Oracle method & REINFORCE & subroutine of getting oracles \\
PSRO iterations & 15 &  number of PSRO iterations  \\
Action Dimension & 500 &  Number of pure strategies available \\
Learning rate & 0.005 & Oracle learning rate \\
Oracle Epochs & 2000 & Oracle total epochs \\
Oracle Epoch Timesteps & 100 & Timesteps per Oracle epoch \\
RAE Gamma & 0.1, 0.5 & Variance aversion parameter \\
Metasolver & RAE SFP & Metasolver method\\
Metasolver Iterations & 100 & Metasolver Iterations \\
\# of seeds & 20 & \# of trials \\ 
\hline \hline

\multicolumn{3}{c}{\textbf{Stag-Hunt Grid-World}}  \\  \hline \hline
Oracle method & MV-PPO \citep{zhang2020mean} & subroutine of getting oracles \\
PSRO iterations & 10 &  number of PSRO iterations  \\
Gore Cost & 2 &  Cost for getting caught by stag \\
PPO Hyperparams & Default SB3 \citep{stable-baselines3} & PPO Hyperparameter values \\
MV-PPO Variance Aversion & 0.15 & PPO Variance Aversion parameter \\
RAE Gamma & 0.15 & Variance aversion parameter \\
Metasolver & RAE SFP & Metasolver method\\
Metasolver Iterations & 100 & Metasolver Iterations \\
\# of seeds & 5 & \# of trials \\ 
\hline \hline

\multicolumn{3}{c}{\textbf{Two-Way Environment}}  \\  \hline \hline
Oracle method & MV-PPO \citep{zhang2020mean} & subroutine of getting oracles \\
PSRO iterations & 7 &  number of PSRO iterations  \\
PPO Hyperparams & Default SB3 \citep{stable-baselines3} & PPO Hyperparameter values \\
MV-PPO Variance Aversion & 0.5 & PPO Variance Aversion parameter \\
RAE Gamma & 0.5 & Variance aversion parameter \\
Metasolver & RAE SFP & Metasolver method\\
Metasolver Iterations & 100 & Metasolver Iterations \\
\# of seeds & 5 & \# of trials \\ 
\hline \hline
\bottomrule
\end{tabular}
}
\end{sc}
\end{table}

\clearpage

\section{Environments}
\label{appendix:envs}
\subsection{Randomly Generated NFGs}
\label{appendix:envs-nfg}
\begin{figure}[ht]
\centering
\includegraphics[width=0.5\linewidth]{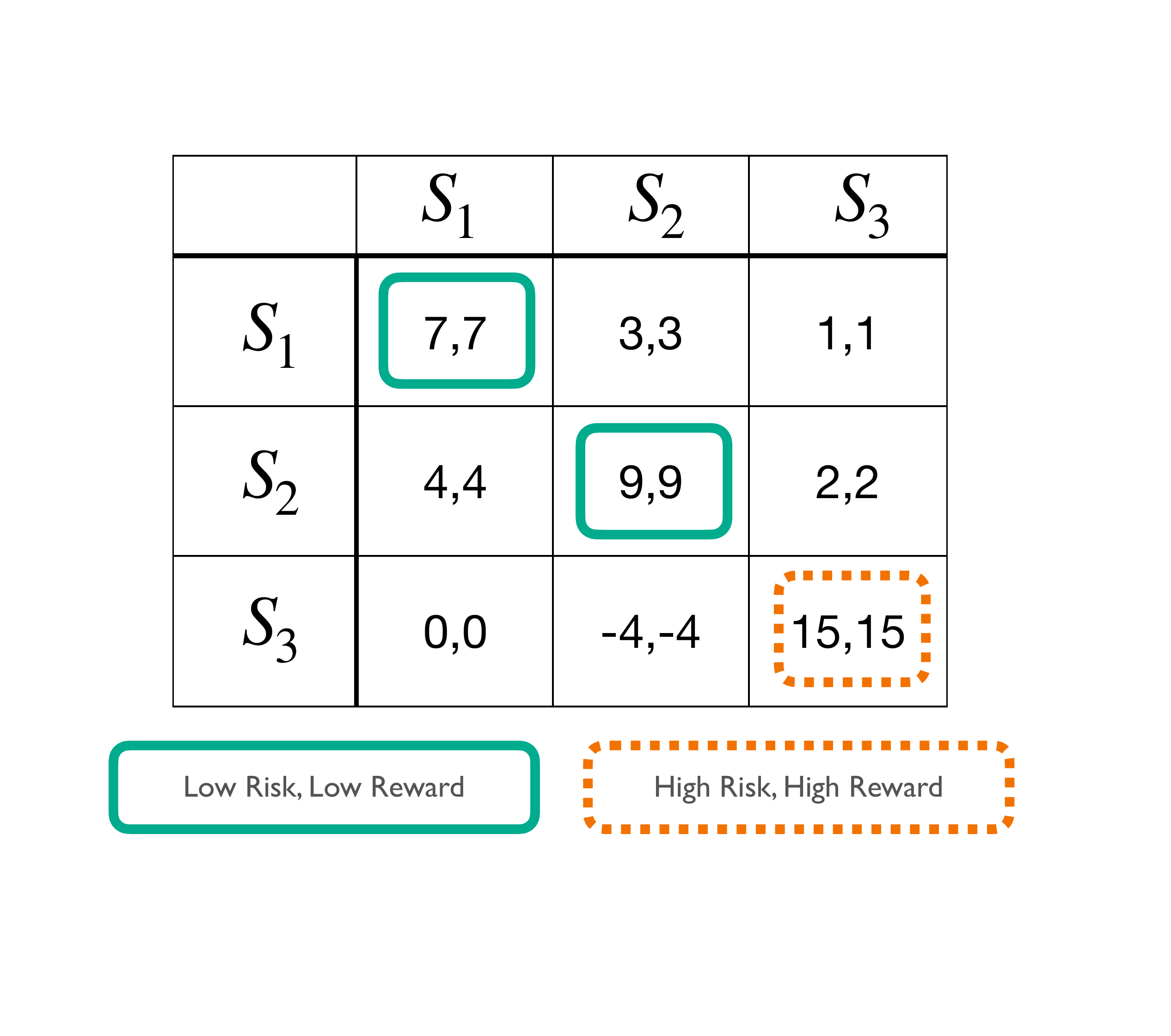}
\caption{An example of a 3 action game. Action $S_3$ (dotted outline) provides a high return assuming successful coordination but high variance in case the opponent does not coordinate correctly.}
\label{fig:coord-game-fig}
\end{figure}

\subsubsection{Characteristics}
\begin{enumerate}
    \item High risk high-reward actions. In these strategies if both players play the same action then they receive a high payoff, however if a player takes a different action then a big negative payoff is received. For example, in Fig. \ref{fig:coord-game-fig}, $S_3$ is the high-risk high-reward strategy.
    \item Low risk low-reward actions. In these strategies if both players play the same action then they receive a lower payoff, however if a player takes a different action then a big smaller payoff is received. For example, in Fig. \ref{fig:coord-game-fig}, $S_1$ and $S_2$ are the low-risk low-reward strategies.
\end{enumerate}

\subsubsection{Generaton}
We randomly generate coordination games with $N$ actions in the following way:

\begin{algorithm}
 \caption{Iterative RAE Generator}
 \begin{algorithmic}[1]
 \STATE \textbf{Initialise:} Empty $N \times N$ payoff matrix $P$
 \STATE \textbf{for} each action $i$ \textbf{do}:
 \STATE \hspace{1em} Sample coordination element, $p_{ii} \sim \mathcal{U}(5,15)$
 \STATE \hspace{1em} Set Payoff matrix element $P_{ii} = |p_{ii}|$
 \STATE \hspace{1em} \textbf{if} $P(X \leq p_{ii}) > 0.9$ \textbf{do}
 \STATE \hspace{2em} \textbf{for} all other actions $j$ \textbf{do}
 \STATE \hspace{3em} Sample anti-coordination element $p_{ij} \sim \mathcal{U}(-10, 15)$
 \STATE \hspace{3em} Set Payoff matrix element $P_{ij} = P_{ji} = p_{ij}$

 \STATE \hspace{1em} \textbf{else do}
  \STATE \hspace{2em} \textbf{for} all other actions $j$ \textbf{do}
 \STATE \hspace{3em} Sample anti-coordination element $p_{ij} \sim \mathcal{U}(0, 10)$
 \STATE \hspace{3em} Set Payoff matrix element $P_{ij} = P_{ji} = p_{ij}$

\STATE \textbf{Return:} $P$. 
\end{algorithmic}
 \end{algorithm}

\clearpage
 
\subsection{Stag Hunt Grid World}
\label{appendix:envs-stag}
Our stag-hunt environment is taken from \citep{prosocial} with minor adjustments to the parameters of the game.

\subsubsection{Characteristics}
\textbf{Details} - 2 Players, 1 Stag, 2 Plants. All spawned in random positions dependent on the seed.

\textbf{State Space} - $5 \times 5$ grid-world. Grid spots are marked as: 0 if nothing on them, 1 if Player is on them, 2 if Stag is on them, 3 if a Plant is on them. Players see the full grid-world with the state space being $\mathcal{S} \in \{0,1,2,3\}^{5 \times 5}$.

\textbf{Action Space} - Actions involve movement in the grid-world in the four cardinal directions. $\mathcal{A} = \{\text{left, right, up, down}\}$.

\textbf{Reward Space} - There are 4 different rewards signals in the game:

\begin{enumerate}
    \item If a Player moves over a Plant they get $r = 2$, and the Plant respawns elsewhere.
    \item If both Players move over the Stag at the same time both receive $r=5$ and the Stag respawns elsewhere.
    \item If a Player moves over the Stag on their own, or the Stag moves over them on their own, the Player receives $r=-2$ and the Stag respawns elsewhere on the grid.
    \item Otherwise $r=0$.
\end{enumerate}

\textbf{The Stag} - At each time-step $t$, the Stag will take one grid-step in the four cardinal directions towards the Player that is closest to it. 

\subsection{Autonomous Driving Environment}
\label{appendix:envs-drive}
Our driving environment is based on the two-way environment from \citep{highway-env} where we make modifications to the reward function to introduce a larger factor of risk-aversion into the game. The goal of the controlled drivers is to reach the end of the road (the destination) whilst avoiding crashing and coming into too close contact with other vehicles. Slow moving drivers populate the roads moving at a constant speed of 20.

\subsubsection{Characteristics}
\textbf{Details} - 2 Players heading opposite directions, 6 other cars on road with even split heading each direction. All spawned in random positions dependent on the seed. 

\textbf{State Space} - For the state space we use the KinematicObservation in \citealt{highway-env}. The KinematicObservation is a $V \times F$ array that describes a list of $V$ nearby vehicles by a set of features of size $F$. We use the default feature set in \citealt{highway-env}, which is the $x$ and $y$ coordinate of the $V$ nearby vehicles and their velocities in the $x$ and $y$ direction. 

\textbf{Action Space} - For the action space we use the DiscreteAction in \citealt{highway-env}. The DiscreteAction is a uniform quantization of the ContinuousAction which allows the agent to directly set the vehicle kinematics, by controlling the throttle $a$ and the steering angle $\delta$.

\textbf{Reward Space} - There are 4 reward signals in the environment:

\begin{enumerate}
    \item If the car crashes $r=-2$.
    \item If the car arrives at the destination $r=2$.
    \item If the car is travelling at a good speed ([25,30]), $r=0.2$.
    \item If the car comes very close to another car $r=-0.1$
    \item Otherwise $r=0$.
\end{enumerate}

\textbf{Other Cars} - Non-Player cars on the road travel at a constant speed of $v=20$ and do not change direction. On each road there are 3 spawned NPC cars ahead of the Player cars.
\clearpage

\section{Baselines}
\label{appendix:baselines}
 For NFG tasks we use the baselines: NE (including risk/dominant payoff NE), THPE and QRE introduced in Sec. \ref{sec:related_work}. For MDP tasks we use the baselines: PSRO-\{Nash, Uniform, Self-Play, THPE, QRE\}. where the brackets refer to the meta-solver used. In the PSRO setting we limit our baselines to algorithms that operate within this framework, and to not consider non-population risk-aversion algorithms (as is standard in PSRO literature). 

 \textbf{THPE and QRE} - For SFP settings, we solve for these equilibria using Fictitious Play by replacing the best-responses with a trembling hand best-response and a Logit Quantal best-response. Empirically, we clarify that the time-average does converge using FP and therefore we do have a THPE and QRE in the games where we used FP (i.e. smaller games)

 For PSRO settings, we treated the new agent best-response for THPE and QRE as vanilla PSRO optimisation agents, i.e. in terms of expected reward. Therefore, it is likely that these baselines are only approximations of the true THPE and QRE, however the lack of a notable solver in these games is an obvious downside of these equilibria.

 \textbf{Nash} - We use vanilla Fictitious Play for any Nash equilibrium solving on NFGs. 

 \textbf{Uniform} - All actions / policies receive equal mixed-strategy probability.

 \textbf{Self-Play} - Only the most recent policy in a populations equals probability in the mixed-strategy, i.e. a pure strategy.
\clearpage

\section{Compute Architecture}
\label{sec:compute}

All experiments run on one machine with:
\begin{itemize}
    \item AMD Ryzen Threadripper 3960X 24 Core
    \item 1 x NVIDIA GeForce RTX 3090
\end{itemize}

\clearpage

\section{Asymmetric Formulations}
\label{appendix:asymmetric}
In the following section we will show the formulation of Sec. \ref{sec:rae} but for the asymmetric case.

\subsection{RAE Derivation}
\label{appendix:asymmetric-utility}

Define the utility for player $i$ of action $a^i_k \in \mathcal{A}^i$ against action $a^j_{k'} \in \mathcal{A}^j$ as $u^i(a^i_k, a^j_{k'})$ and the full utility matrix as $\tM^i$, where the entry $\tM^i_{k,k'}$ refers to $u^i(a^i_k, a^j_{k'})$ and $\tM^i_k$ refers to $u^i(a^i_k, a^j_{k'}) \; \forall k' $, i.e. the vector of utilities that action $a^i_k$ receives against all other actions. We now define the \textit{expected} utility of the mixed-strategy for player 1 $\boldsymbol{\sigma}^i$ versus the mixed strategy for player 2 $\boldsymbol{\varsigma}^j$ as

\begin{equation}
\begin{aligned}
    u^i(\boldsymbol{\sigma}^i, \boldsymbol{\varsigma}^j, \tM^i) &= \sum_{a^i_k \in A^i} \sum_{a^j_{k'} \in A^j} \sigma(a^i_k) \varsigma(a^j_{k'}) u^i(a^i_k, a^j_{k'}) \\ &= \boldsymbol{\sigma}^{i, T}\cdot \tM^i \cdot \boldsymbol{\varsigma}^j.
\end{aligned}
\end{equation}

The weighted co-variance matrix for $\tM^i$ (i.e. the variance of utility values) is a $|A^i| \times |A^i|$ matrix $\boldsymbol{\Sigma}_{\tM^i, \boldsymbol{\varsigma}^j}$ = $[c^i_{k,k'}]$ with entries 
\begin{equation}
\begin{aligned}
c^i_{k,k'} = \sum_{a^j_d \in A^j}  \varsigma^j(a^j_{d}) \big(u^i(a^i_d,a^j_{k})-\bar{\tM}^i_{k}\big)\big(u^i(a^i_d, a^j_{k'}) - \bar{\tM}^i_{k'}\big),
\end{aligned}
\end{equation}
where $\bar{\tM}^i_{k} = \sum_{k'=1}^{|A^j|} \varsigma(a^j_{k'}) u(a^i_k, a^j_{k'})$ is the EU for Player $i$'s action $k$ given the opponent mixed-strategy $\boldsymbol{\varsigma}^j$. This allows us to define the mixed-strategy $\boldsymbol{\sigma}^i$ variance utility as:
\begin{equation}
\begin{aligned}
\operatorname{Var}(\boldsymbol{\sigma}^i, \boldsymbol{\varsigma}^j, \tM^i) &= \sum_{k=1}^{|A^i|} \sum_{k'=1}^{|A^j|} \sigma(a^i_k) \sigma(a^j_{k'}) c_{k, k'} \\ 
&= \boldsymbol{\sigma}^{i, T} \cdot \boldsymbol{\Sigma}_{\tM^i, \boldsymbol{\varsigma}^j} \cdot \boldsymbol{\sigma}^i.
\end{aligned}
\end{equation}

The final utility function $r^i$ which considers expected and variance utility for mixed-strategy $\boldsymbol{\sigma}^i$ is,
\begin{align}
    r(\boldsymbol{\sigma}^i, \boldsymbol{\varsigma}^j, \tM^i) = u^i(\boldsymbol{\sigma}^i, \boldsymbol{\varsigma}^j, \tM^i) - \gamma^i \operatorname{Var}(\boldsymbol{\sigma}^i, \boldsymbol{\varsigma}^j, \tM^i),
\end{align}
where $\gamma^i \in \mathbb{R}$ is the risk-aversion parameter. 

In order to define RAE, we first define the best-response map:
\begin{equation} 
\begin{aligned}\label{eq:br-map-asymmetric}
\boldsymbol{\sigma}^{*,}(\boldsymbol{\varsigma}^j) \in \argmax_{\boldsymbol{\sigma}^i}r^i(\boldsymbol{\sigma}^i, \boldsymbol{\varsigma}^j, \tM^i) \\
\text{s.t. } \sigma(a) \geq \epsilon  \text{  }, \forall a \in A^i\\
\boldsymbol{\sigma}^{i, T}\textbf{1} = 1, 
\end{aligned}
\end{equation}

\begin{definition}[RAE]\label{appendix:rae_def} A strategy profile $\{\boldsymbol{\sigma}^i, \boldsymbol{\varsigma}^j\} $ is a risk-averse equilibrium if both $\boldsymbol{\sigma}^i$ and $\boldsymbol{\varsigma}^j$ are risk-averse best responses, in that they satisfy Eq. \ref{eq:br-map-asymmetric}, to each other.
\end{definition}

\subsection{Multi-population PSRO-RAE}
\label{appendix:asymmetric-psro}
\subsubsection{PSRO Outer Loop}

 At every iteration $t \leq T$, a \textit{player} $i$ is defined by a population of fixed \textit{agents} $\boldsymbol{\Phi}^i_{t} = \boldsymbol{\Phi}^i_{0} \cup \left\{\boldsymbol{\phi}^i_1, \boldsymbol{\phi}^i_2, ... , \boldsymbol{\phi}^i_{t}\right \}$, where $\boldsymbol{\Phi}^i_{0}$ is the initial random agent.  

\subsubsection{PSRO Inner Loop}

\textbf{a, d) Meta-Game \& Covariance Matrix}
At the start of the iteration $t$ inner loop, each player $i$ has population with a \textit{meta-game} $\tM^i_t$, an EU matrix between all the \textit{agents} in its own population and that of the opponent $j$, with individual entries $M^i(\boldsymbol{\phi}^i_k, \boldsymbol{\phi}^j_{k'}) \; \forall \boldsymbol{\phi}^i_k \in \boldsymbol{\Phi}^i_{t}, \boldsymbol{\phi}^j_{k'} \in \boldsymbol{\Phi}^j_{t}.$ In addition, each player $i$ with population also generates a covariance matrix $\mathbf{\Sigma}_{\tM^i_t}$ defined by Eq. \ref{eq:covariance}. At the end of iteration $t$ inner loop, both $\tM^i_t$ and $\mathbf{\Sigma}_{\tM^i_t}$ are updated to include a new agent. 

\textbf{b) Meta Distribution}
To use $\boldsymbol{\Phi}^i_t$ we require a way to select which $\phi^i_t \in \boldsymbol{\Phi}^i_t$ will be used as training opponents. The function $f^i$ is a mapping $f^i : \tM^i_t \rightarrow [0,1]^t$ which takes as input a meta-game $\tM^i_t$ and outputs a \textit{meta-distribution} $\boldsymbol{\sigma}^i_t = f^i(\tM^i_t)$. The output $\boldsymbol{\sigma}^i_t$ is a probability assignment to each \textit{agent} in the population $\boldsymbol{\Phi}^i_t$ which is the equivalent of a mixed-strategy in a NFG, except actions are now RL policies. We apply RAE (Def. \ref{rae_def}) as the meta-solver. As $\boldsymbol{\phi}^i$ are RL policies then the policies are sampled by their respective probability in $\boldsymbol{\sigma}^i_t$.

\textbf{c) Best Response Oracle}
At each epoch $\boldsymbol{\Phi}^i_t$ is augmented with a new agent that is a \textit{best-response} (BR) to the meta-distribution $\boldsymbol{\sigma}^i_t$. When selecting the BR oracle one aims to optimise the same objective function of that optimised by the meta-distribution. For example, in Vanilla PSRO the Nash meta-distribution optimises for environment reward and the BR oracle also optimises a new agent in terms of environment reward. This can be found with any optimisation process such as RL or an evolutionary algorithm. In our setting the meta-distribution optimises two metrics, the EU and the variance utility and therefore we similarly need a BR oracles that optimises the same dual objective. 

In terms of RL quantities, this translates to maximising the expected total-RL reward (i.e. the total of the per-step rewards) whilst minimising the variance of the total-RL reward caused by the sampling of different RL agents from $\boldsymbol{\sigma}^i_t$. 

To achieve this, we follow the approach of \citep{zhang2020mean} that optimises both the total RL-reward and per-step RL-reward variance by solving an augmented MDP where the per-step reward $g^i_t$ is replaced by:
\begin{align*}
    \hat{g}^i_t = g^i_t - \lambda (g_t^i)^2 + (2 \lambda g_t^i y_i)
\end{align*}
where $y_i = \frac{1}{T}\sum_{t=1}^T g_t^i$ is the average of the per-step rewards during the data collection phase. 

We choose the per-step RL reward as it is an upper bound of the total-RL reward variance \citep{bisi2019risk}, therefore reducing per-step variance will also reduce total variance, and is more effective computationally \citep{zhang2020mean}. Additionally, as this variance is also with respect to the sampling probability defined by $\boldsymbol{\sigma}^i_t$ this optimises the correct co-variance matrix which is also weighted by $\boldsymbol{\sigma}^i_t$. 
\clearpage

\section{QRE Failure Case}
In the following section we present results on the two-action driving game described in Sec. 1 of the main article and displayed in Fig. 5. 

\begin{figure*}[h!]
\centering
\includegraphics[width=1.0\linewidth]{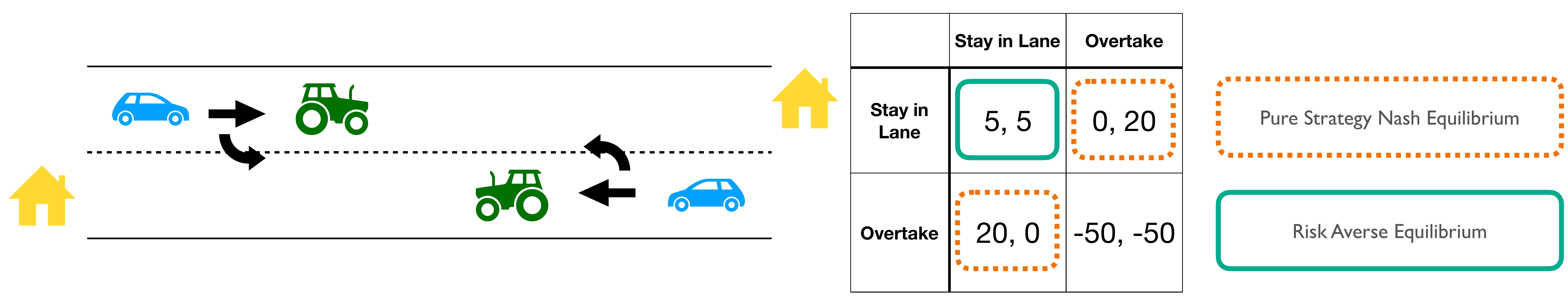}
\caption{Two-action driving risk game.}
\label{fig:self-drive-appendix}
\end{figure*}

We specifically utilise this game to show a failure case of QRE as a risk-sensitive solution. Ideally, a risk-sensitive solution concept would only play the Stay in Lane strategy as the Overtake strategy has far too high potential downside risk. 

\begin{figure}[h!]
\centering
\includegraphics[width=1.0\linewidth]{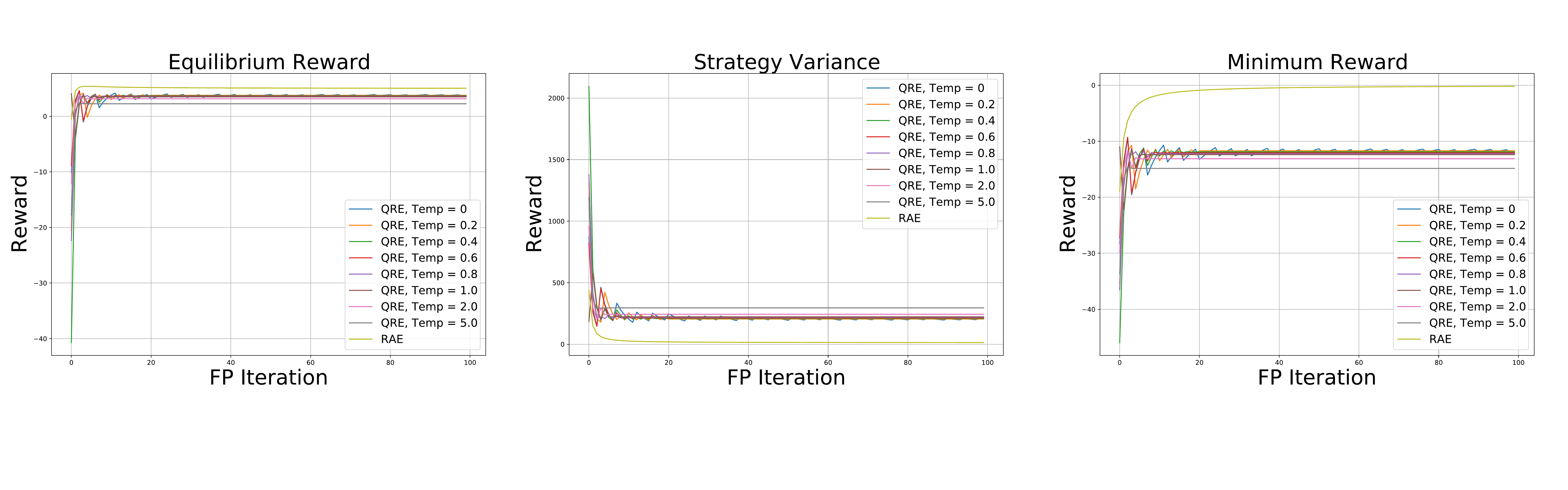}
\caption{QRE and RAE results on two-action driving game.}
\end{figure}

As can be seen from the results in Fig. 6, for a large sample of QRE hyperparameters the equilibrium found is high variance with potential poor downside performance. We believe this is because the very large costs of the errors are easily picked up by variance analysis, but not so easily by the setup of QRE.

\clearpage

\section{Variance vs. Standard Deviation}
It is worth noting that a property of variance is that it is not scale-invariant with relation to the utilities of the game, and one might suggest using standard deviation (STD) instead as it is scale-invariant. However, we choose to stick with variance due to its mathematical properties, notably that because it is quadratic with respect to the strategy probabilities $\boldsymbol{\sigma}$ QP can be used, whereas STD is not quadratic w.r.t $\boldsymbol{\sigma}$ and QP cannot be used.


\end{document}